\documentclass[preprint,11pt,nonatbib]{elsarticle}
\usepackage[margin=1in]{geometry}
\usepackage{graphicx}
\graphicspath{{Paper/Materials/Images/}{Materials/Images/}{Images/}}
\usepackage{amsmath,amssymb}
\usepackage{caption}
\usepackage{subcaption}
\usepackage{float}
\usepackage{booktabs}
\usepackage{multirow}
\usepackage{setspace}
\usepackage{enumitem}
\usepackage[hidelinks]{hyperref}
\usepackage{url}
\usepackage{newunicodechar}
\newunicodechar{−}{-}
\usepackage{tikz}
\usetikzlibrary{arrows.meta, positioning, shadows, calc}
\usepackage{tabularx}
\usepackage{csquotes}
\makeatletter
\let\c@author\relax
\makeatother
\usepackage[
  backend=bibtex,
  style=authoryear,
  maxcitenames=2,
  maxbibnames=99
]{biblatex}
\begin{document}
\begin{frontmatter}
\title{Predicting \textit{Pseudo-nitzschia} harmful algal blooms along the Portuguese Coast using satellite-derived predictors}
\author[aff1]{Ayman Bnoussaad\corref{cor1}}
\ead{ayman.bnoussaad@tecnico.ulisboa.pt}
\cortext[cor1]{Corresponding author.}
\author[aff2]{El Khalil Cherif}
\ead{el.k.cherif@tecnico.ulisboa.pt}
\author[aff2]{Ligia Pinto}
\ead{ligia.pinto@tecnico.ulisboa.pt}
\author[aff2]{Ramiro Neves}
\ead{ramiro.neves@tecnico.ulisboa.pt}
\author[aff3,aff4]{Alexandra D. Silva}
\ead{amsilva@ipma.pt}
\author[aff2]{Alexandre Bernardino}
\ead{alexandre.bernardino@tecnico.ulisboa.pt}
\affiliation[aff1]{organization={ISR -- Institute for Systems and Robotics, Lisbon, Portugal.}}
\affiliation[aff2]{organization={MARETEC/LARSyS, Associação do Instituto Superior Técnico para a Investigação e Desenvolvimento, Lisboa, Portugal.}}
\affiliation[aff3]{organization={IPMA -- Portuguese Institute for Sea and Atmosphere, Algés, Portugal.}}
\affiliation[aff4]{organization={CIIMAR -- Centro Interdisciplinar de Investigação Marinha e Ambiental / CIMAR LA, Matosinhos, Portugal.}}
\begin{abstract}
\textit{Pseudo-nitzschia} diatoms pose recurrent risks to coastal ecosystems and shellfish harvesting along the Portuguese Atlantic coast, where operational forecasting of bloom occurrence can support monitoring programmes by enabling earlier and targeted management responses. Here we develop and evaluate a spatio-temporal machine-learning framework to predict harmful algal bloom (HAB) occurrence using exclusively satellite-derived predictors under realistic forecasting constraints. We first characterised environmental and biological variability across the shellfish production zones (L1--L9) along the Portuguese coast using 5,882 observations, providing a system-wide context. Predictive models were then developed for the core study region (zones L1--L2), a hotspot for \textit{Pseudo-nitzschia} and domoic acid events, using a decade-long matched dataset (2013--2023; 1{,}440 observations and more than 1{,}000 satellite-based predictors, including sea surface temperature, an empirically derived upwelling index, chlorophyll-\textit{a}, and plankton functional types). To account for coastal heterogeneity, sampling locations were partitioned into ecologically meaningful sub-regions using a river-aware spatial clustering scheme. A stringent spatio-temporal cross-validation strategy that simultaneously withholds entire years and spatial clusters prevents temporal and spatial leakage and closely mimics real-world forecasting conditions. Across multiple model classes and feature configurations, HAB occurrence proved moderately predictable. Ensemble tree-based methods achieved the strongest discrimination: Random Forest reached $0.74 \pm 0.05$ with environmental predictors; Extra Trees reached $0.77 \pm 0.06$ with biological variables added. Feature-importance analyses revealed that seasonal structure, spatial context, and lagged environmental conditions dominate model decisions, while biological indicators refine bloom likelihood within physically favourable periods. The framework demonstrates operationally relevant skill for satellite-supported HAB early-warning systems along eastern boundary upwelling coasts.
\end{abstract}
\begin{keyword}
\textit{Pseudo-nitzschia} \sep Harmful algal blooms \sep Remote sensing \sep Machine learning \sep Spatio-temporal forecasting \sep Portuguese Atlantic coast
\end{keyword}
\end{frontmatter}
\section*{Highlights}
\begin{itemize}[noitemsep,topsep=2pt,leftmargin=*]
  \item Spatio-temporal ML predicts \textit{Pseudo-nitzschia} HABs from satellite data alone.
  \item Strict year $\times$ cluster cross-validation prevents temporal and spatial leakage.
  \item Extra Trees with biological predictors reaches ROC--AUC $0.77 \pm 0.06$.
  \item Lagged SST, upwelling and chlorophyll-\textit{a} are the dominant predictors.
  \item Framework supports operational early warning on the Iberian upwelling margin.
\end{itemize}

\section{Introduction}
The diatom genus \textit{Pseudo-nitzschia} is globally distributed in coastal and oceanic waters, though not all species or bloom events are harmful. Certain species produce the neurotoxin domoic acid, the causative agent of Amnesic Shellfish Poisoning, whose consequences range from severe neurological illness in humans to mass mortality events in marine mammals and seabirds \parencite{silva2016hab, sun2011effects}. The transition from a harmless presence to a toxic bloom is often driven by environmental stressors including nutrient shifts, temperature fluctuations and upwelling dynamics, which can trigger toxin production rapidly and at relatively low cell abundances \parencite{howard2007nitrogen, gobler2020climate}. Given the increasing frequency of these events and their well-documented impacts on public health, fisheries and aquaculture, there is a clear need for predictive tools that can anticipate bloom conditions and support earlier and more targeted management responses \parencite{griffith2020harmful, hallegraeff2010ocean}.
Routine monitoring of \textit{Pseudo-nitzschia} relies on \textit{in situ} sampling, which remains the operational standard for verifying cell concentrations and biotoxin levels in shellfish. However, the spatial and temporal coverage of \textit{in situ} programmes is inherently limited by cost and logistics, and broad-scale satellite observations offer a complementary source of information on the physical and biological conditions that precede bloom development \parencite{cherif2021comparison}. Standard ocean colour products cannot distinguish \textit{Pseudo-nitzschia} from other phytoplankton, but statistical and machine learning approaches applied to satellite-derived environmental and biological variables have shown meaningful skill in predicting bloom-conducive conditions in several coastal systems \parencite{blondeau2021review, gonzalez2024new}. The California Harmful Algae Risk Mapping system demonstrated that combining satellite observations with physical model outputs can support daily operational forecasting of elevated \textit{Pseudo-nitzschia} abundance \parencite{anderson2016initial}, illustrating the potential of this approach for routine monitoring support. Realising that potential in other regions requires frameworks that are adapted to local physical forcing, grounded in long-term observational records, and evaluated under realistic forecasting constraints.
The Portuguese Atlantic coast is a particularly compelling setting for this purpose. It lies within the Iberian eastern boundary upwelling system, where recurrent wind-driven upwelling and riverine inputs create highly dynamic coastal conditions that are known to favour episodic \textit{Pseudo-nitzschia} proliferation \parencite{pitcher2010physical, palma2010can}. Portugal supports a nationally important bivalve aquaculture industry, and shellfish harvesting closures due to biotoxins carry significant socioeconomic consequences \parencite{braga2023bivalve}. Operational monitoring is conducted by the Instituto Português do Mar e da Atmosfera (IPMA), which manages classified shellfish production areas along the coast and maintains long-term weekly to monthly records of phytoplankton abundance and shellfish biotoxins. The northern production zones L1 and L2 have been repeatedly identified as a hotspot for recurrent \textit{Pseudo-nitzschia} blooms and domoic acid events \parencite{torres2019pseudo, trainer2012pseudonitzschia}, and over the 2013--2023 study period accumulated the highest cumulative number of harvesting ban days along the Portuguese coast. These long-term, spatially resolved datasets, together with well-characterised physical forcing, make the Portuguese coast an ideal setting for developing and validating predictive models that translate satellite observations into actionable risk assessments \parencite{silva2016hab, torres2019pseudo}. Despite routine monitoring, management decisions currently rely largely on historical patterns and seasonal experience, and cell concentration alone is an imperfect indicator of public health risk since harmful toxin levels can occur at relatively low cell abundances \parencite{trainer2012pseudonitzschia, torres2019pseudo}. Understanding the environmental drivers of bloom development therefore remains central to improving both the efficiency and the early-warning capability of existing monitoring programmes.

These considerations motivate the data-driven forecasting framework developed in this study. We integrate satellite-derived physical and biological predictors; sea surface temperature, an empirically derived upwelling index, chlorophyll-\textit{a} and plankton functional types, together with their lagged values, with long-term \textit{in situ} records of \textit{Pseudo-nitzschia} cell concentrations from IPMA to predict bloom occurrence along the northern Portuguese coast. The framework is designed explicitly for operational realism: predictors are derived exclusively from satellite observations, the supervised target is defined from \textit{in situ} measurements independently of the predictors, and model evaluation employs a stringent spatio-temporal cross-validation protocol that prevents both temporal and spatial leakage. The following section describes the study region, data sources, modelling strategy and evaluation metrics used to develop and validate this predictive framework.
\section{Methodology}
The overarching objective of this study is to quantify the operational predictability of \textit{Pseudo-nitzschia} (PN) harmful algal bloom (HAB) occurrence along the Portuguese Atlantic coast using exclusively satellite-derived predictors under physically consistent evaluation constraints. To achieve this, we (i) construct a unified spatio-temporal matchup dataset that collocates \textit{in situ} PN observations with daily satellite-derived environmental (SST, UI) and biological (CHL, PFTs) variables, including lagged conditions representative of information availability under near-real-time monitoring; (ii) define an operational HAB-occurrence target from PN cell concentrations and evaluate predictive skill under strong class imbalance; (iii) implement an ecologically informed spatial partitioning based on river influence (ROFI) and a stringent spatio-temporal cross-validation protocol that prevents both temporal and spatial leakage through the simultaneous withholding of calendar years and spatial clusters via union-of-year-and-cluster blocking; (iv) compare forecasting skill across progressively enriched feature configurations (seasonal+spatial, seasonal+spatial+environmental, seasonal+spatial+environmental+biological) and multiple model classes (linear, margin-based, and tree-based ensembles); and (v) interpret model outputs in a decision-support context through threshold-dependent trade-off analysis and spatially explicit risk mapping to support adaptive monitoring strategies.
\subsection{Location and Geographical Features}
\label{subsec:location_geo}
\begin{figure}[H]
\centering
\includegraphics[width=0.55\textwidth]{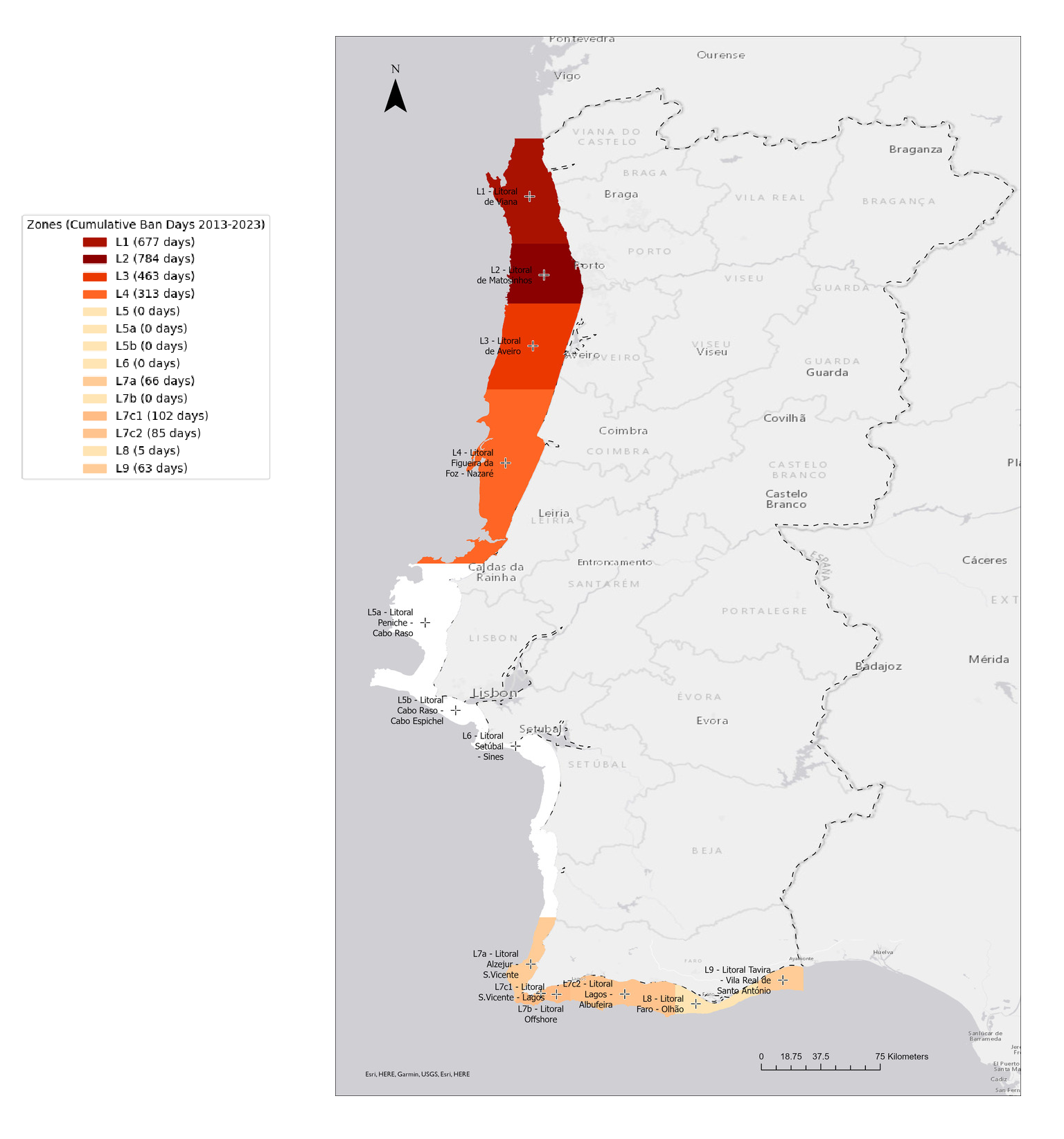}
\caption{Cumulative shellfish harvesting ban days reported by IPMA for each production area between 2013 and 2023.}
\label{fig:tox_bans_map}
\end{figure}

\begin{figure}[H]
\centering
\includegraphics[width=\textwidth]{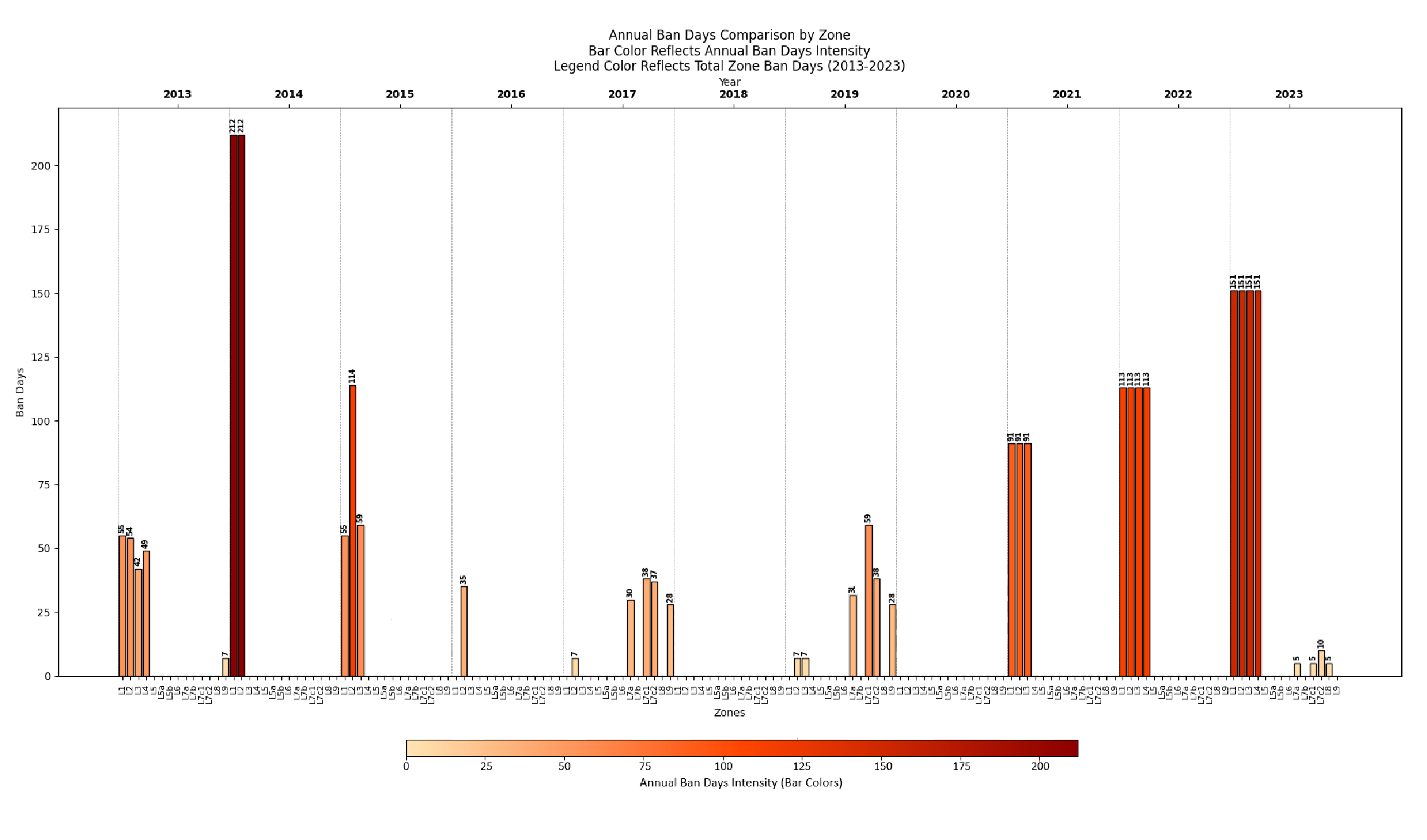}
\caption{Annual shellfish harvesting ban days by production area (2013--2023). Bar colours represent total ban days per zone.}
\label{fig:tox_bans_hist}
\end{figure}

This study was conducted along the Portuguese Atlantic coast, a dynamically forced eastern boundary region characterised by strong latitudinal gradients in oceanography, biogeochemistry, and anthropogenic pressure. The coastline extends from the Minho River in the northwest, bordering Spain, to Vila Real de Santo António in the southeast, adjacent to the Gulf of Cádiz. This Iberian coastal margin supports nationally important fisheries and aquaculture activities and is subject to recurrent harmful algal bloom (HAB) events that require routine environmental monitoring and management interventions. The interaction between wind-driven upwelling, riverine nutrient inputs, complex shelf circulation, and human activity makes the Portuguese coast a particularly suitable natural laboratory for studying and forecasting PN dynamics.
Operational monitoring of phytoplankton and marine biotoxins in Portugal is conducted by IPMA, which defines the classified shellfish production areas and manages harvesting restrictions based on public-health risk within these zones; thirteen officially classified shellfish production zones, in addition to estuarine zones. These production zones constitute the spatial units used to regulate shellfish harvesting, issue biotoxin alerts, and implement public-health advisories. For descriptive purposes, the Portuguese coast was grouped into three broad regions: four zones in the Northern region (L1--L4), three in the Central region (L5--L6, with L5 subdivided into L5a and L5b), and six in the Southern region (L7--L9, with further subdivisions of L7 and L7c). Each zone is defined by fixed geographic boundaries and typically extends offshore to approximately the 100~m isobath, reflecting both broadly consistent ecological features and practical considerations for fisheries management.
Within this national context, the northern Portuguese coast, specifically regions L1 and L2, have been repeatedly identified as hotspots for \textit{Pseudo-nitzschia} blooms and associated domoic acid (DA) events \parencite{torres2019pseudo, trainer2012pseudonitzschia}. These zones were therefore selected as the core modelling region in this study, specifically due to their long and consistent monitoring records, high frequency of HAB occurrences, and relatively well-defined physical forcing, which together support robust model training and evaluation while providing a framework that can be extended to other coastal sectors in future applications. Over the 2013--2023 study period, zones L1 and L2 experienced the highest cumulative number of harvesting ban days (L1: 677 days; L2: 784 days), reflecting their recurrent exposure to DA--related management actions. Although toxicity prediction is not the primary modelling target in this study, the historical prevalence of bans on harvesting highlights the ecological and socio-economic relevance of this region and further motivates its selection as the core modelling region for this predictive framework (Figures~\ref{fig:tox_bans_map} and~\ref{fig:tox_bans_hist}).

This region is influenced by a dense network of rivers, including the Minho, Lima, Cávado, and Douro, which play a central role in shaping coastal environmental conditions. River discharge generates nearshore plumes that shape gradients in stratification, turbidity, nutrient availability, and phytoplankton biomass, collectively referred to as \emph{Regions of Freshwater Influence} (ROFIs). These plumes typically form buoyancy-driven surface layers that spread offshore and alongshore from river mouths, generating strong horizontal gradients in stratification and nutrient availability that structure coastal ecological regimes \parencite{hornerdevine2015plumes}. River-driven regimes generally decay offshore and alongshore with distance from river mouths and exert strong control on coastal phytoplankton dynamics across river-dominated shelves \parencite{kristiansen1997phytoplankton2,simionato2006rioplata,osadchiev2022river}. Satellite observations and coupled physical--biogeochemical modelling studies further demonstrate that river plumes form coherent spatial regimes that can be detected remotely and used to delineate coastal subregions sharing common environmental forcing \parencite{dodrill2022columbia}. This provides the physical basis for the river-aware spatial clustering approach adopted here, described in full in \ref{app:rofi_clustering}.
\begin{figure}[H]
\centering
\includegraphics[width=0.9\textwidth]{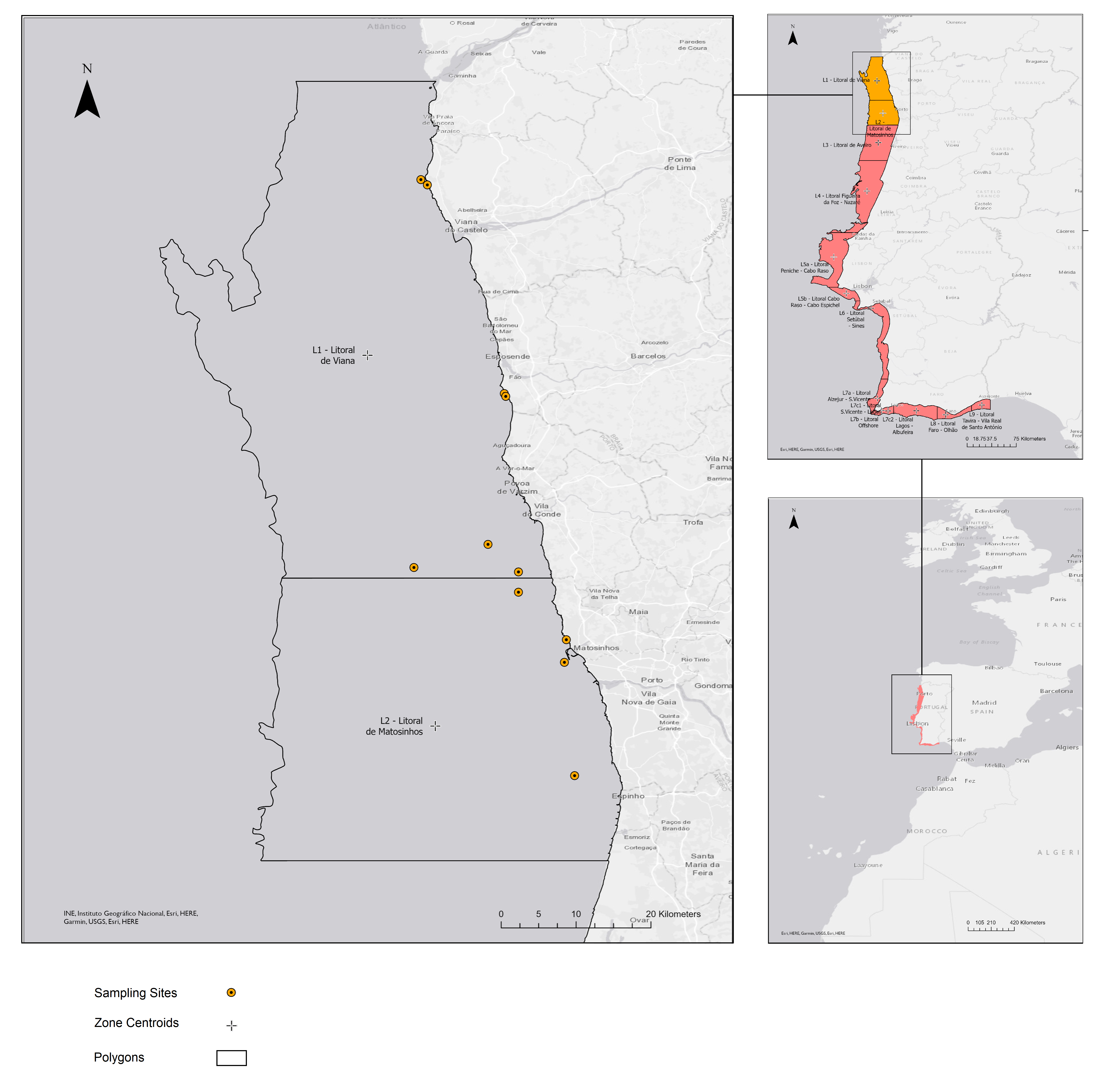}
\vspace{0.9em}
\caption{Spatial distribution of the shellfish production areas along the Portuguese Atlantic coast. Areas L1 (Litoral de Viana) and L2 (Litoral de Matosinhos) are highlighted together with the sampling locations used in this study. These areas are the primary modelling domain for the predictive framework developed in this study.}
\label{fig:fig1}
\end{figure}
In addition to riverine inputs, the northern Portuguese coast is strongly influenced by seasonal wind-driven upwelling associated with persistent northerly winds during spring and summer. These upwelling events inject nutrient-rich subsurface waters onto the continental shelf, fostering high primary productivity and recurrent phytoplankton blooms \parencite{cembella2023emerging}. The interaction between episodic upwelling, nearshore retention, and riverine freshwater inputs creates a highly dynamic coastal environment characterised by strong gradients in stratification, nutrient availability, and circulation. Such conditions are widely recognised as conducive to episodic \textit{Pseudo-nitzschia} (PN) proliferation, reinforcing the designation of this region as a priority target for HAB forecasting and management. The intensity and spatial variability of wind-driven coastal upwelling along the Portuguese margin are illustrated in Figures~\ref{fig:UI_map} and~\ref{fig:UI_hist}, which show the spatial distribution and zonal range of the mean Downwelling/Upwelling Index across coastal zones for the period 2020--2023.
Superimposed on these physical drivers, the region is also subject to significant anthropogenic pressures associated with urban and industrial activity, which can enhance the delivery of nutrients and trace metals to coastal waters and modulate phytoplankton growth, species composition, and physiological stress responses \parencite{reis2017trace,cembella2023emerging}. Regional contaminant assessments derived from the EMODnet Marine Chemistry aggregated datasets (v2024) further indicate spatial gradients in trace-metal concentrations along the Portuguese coast. Median cadmium and lead concentrations in biota are comparatively higher in northern coastal waters, whereas nickel concentrations tend to be higher in southern regions. Although these patterns do not imply exceedance of regulatory thresholds, they highlight regional differences in hydrological forcing and anthropogenic influence that may shape coastal biogeochemical conditions relevant to HAB dynamics
relevant to HAB dynamics\footnote{\label{fn:emodnet}European Marine Observation and Data Network (EMODnet) (2024), ``Metals and metalloids in biota dataset,'' \href{https://emodnet.ec.europa.eu/geonetwork/srv/eng/catalog.search}{link}.} (Figure~\ref{fig:trace_metals}).
\begin{figure}[H]
\centering
\includegraphics[width=1.0\textwidth]{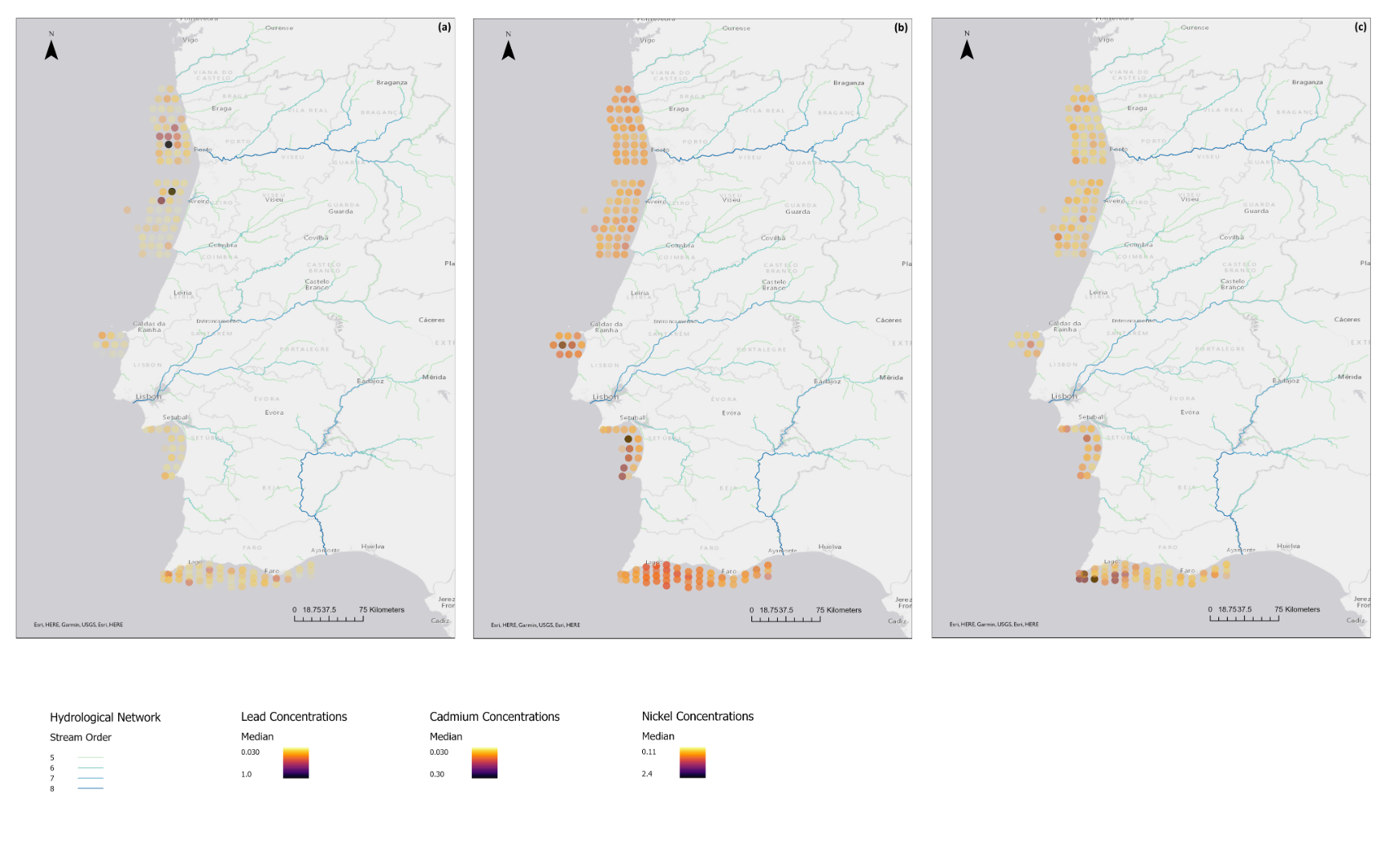}
\vspace{0.9em}
\caption{Spatial distribution of trace metals and metalloids along the Portuguese coast derived from EMODnet marine chemistry datasets (v2024). Panels show median concentrations since 2012 for (a) Lead, (b) Cadmium, and (c) Nickel. Northern coastal waters exhibit elevated lead and cadmium concentrations relative to the rest of the Portuguese shelf, whereas nickel concentrations are comparatively higher in southern regions.}
\label{fig:trace_metals}
\end{figure}

\begin{figure}[H]
\centering
\includegraphics[width=0.35\textwidth]{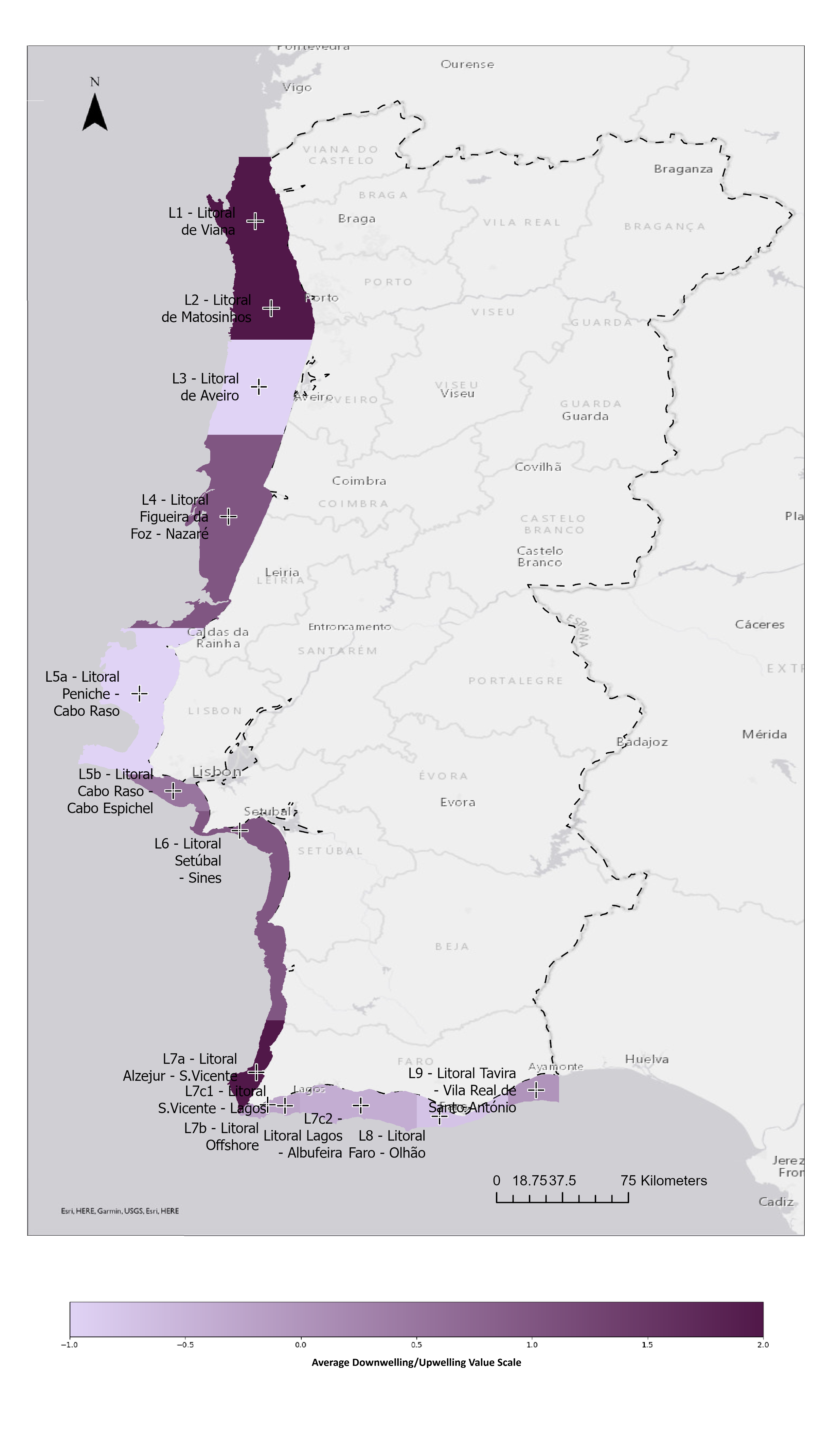}
\caption{Mean coastal Downwelling/Upwelling Index by production area along the Portuguese margin (2020--2023). Map shading represents mean intensity per coastal area (units: m$^{3}$\,s$^{-1}$\,km$^{-1}$). The northern coast exhibits the strongest upwelling regime.}
\label{fig:UI_map}
\end{figure}

\begin{figure}[H]
\centering
\includegraphics[width=0.9\textwidth]{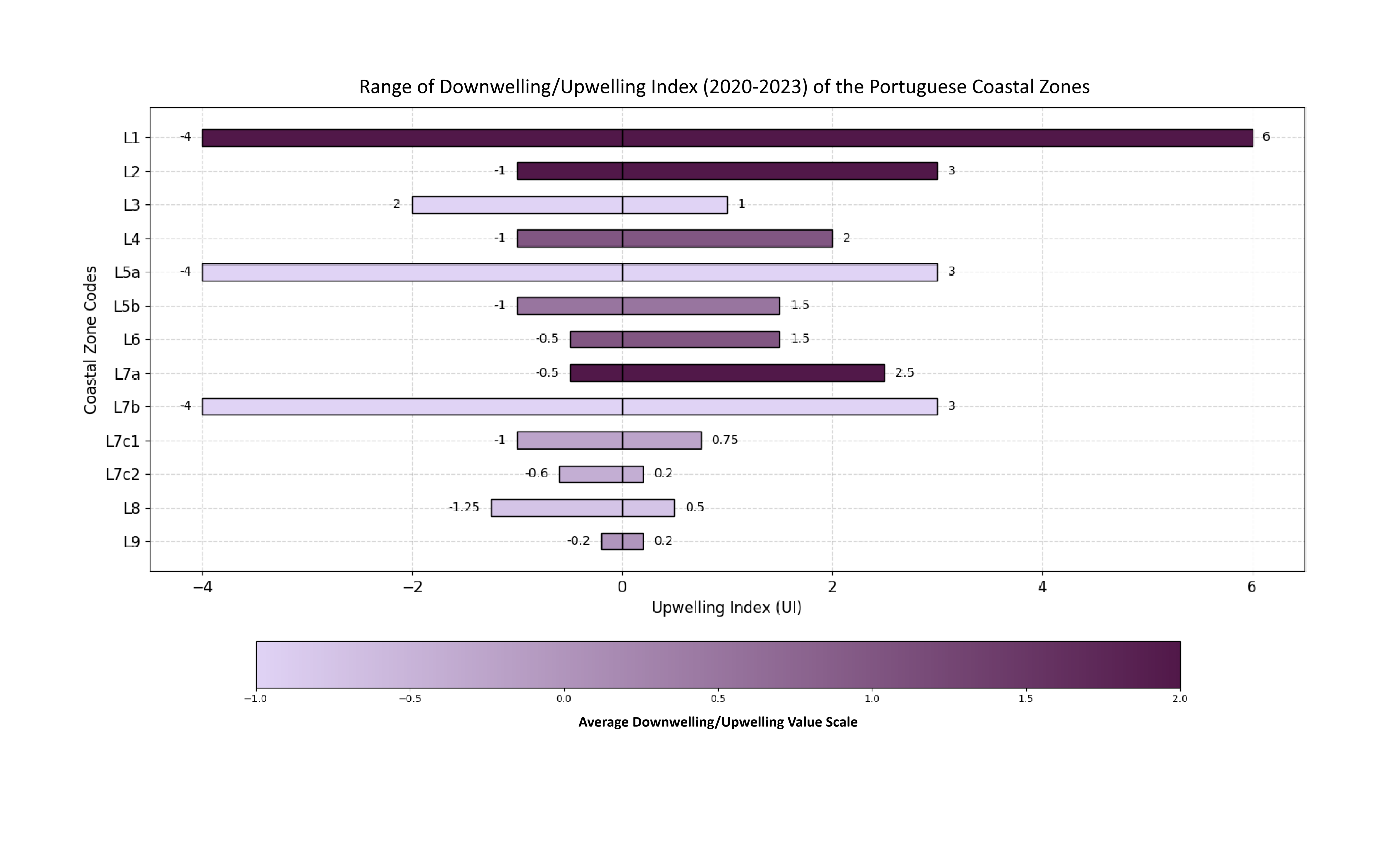}
\caption{Zonal ranges of the coastal Downwelling/Upwelling Index along the Portuguese margin (2020--2023). Bar colours represent mean upwelling/downwelling intensity for each coastal area (units: m$^{3}$\,s$^{-1}$\,km$^{-1}$).}
\label{fig:UI_hist}
\end{figure}

\begin{figure}[H]
\centering
\includegraphics[width=0.55\textwidth]{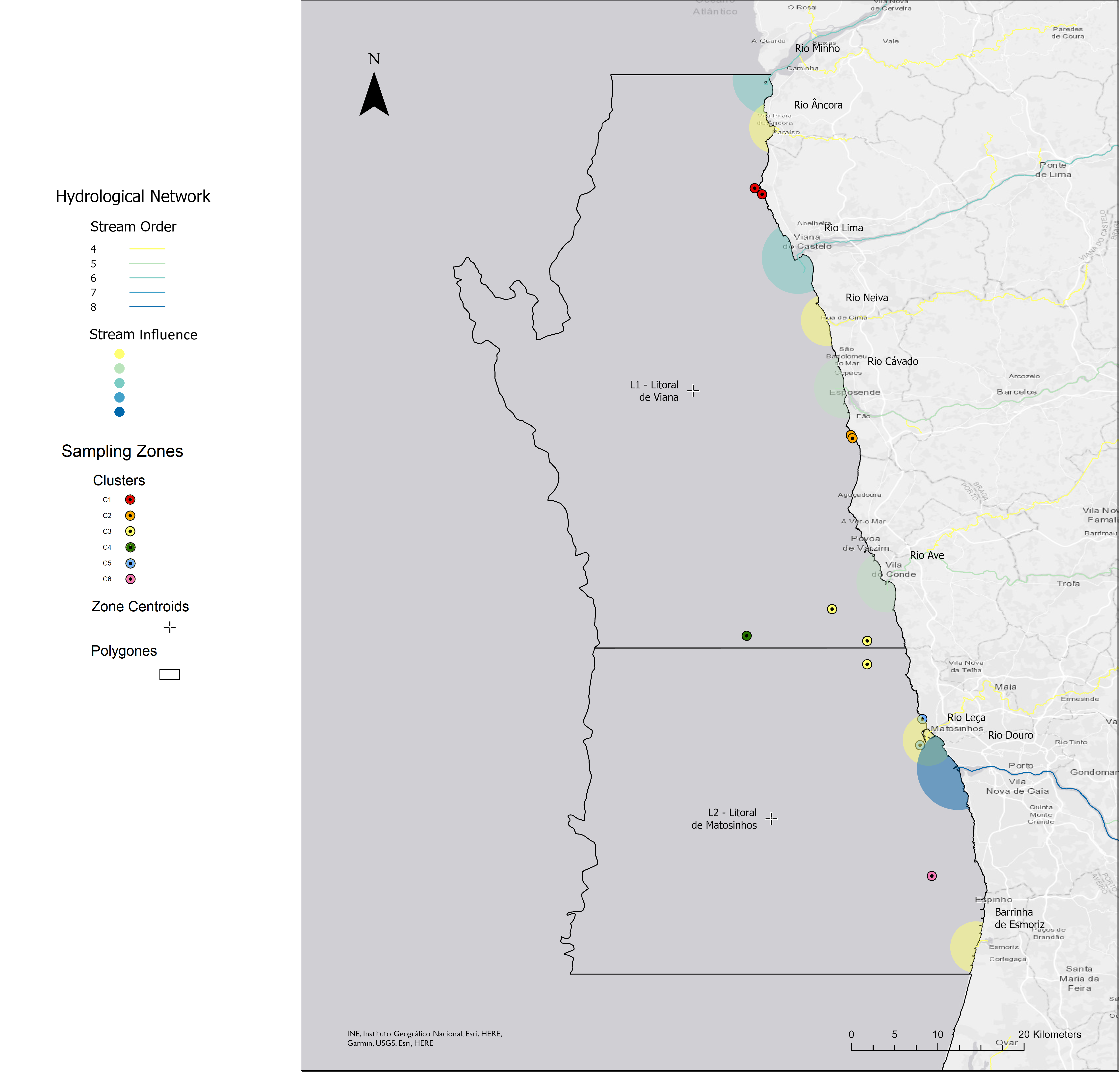}
\caption{River-aware spatial clustering of sampling locations along the northern Portuguese coast. River mouths and hydrological networks are shown and coloured by stream order, representing the relative magnitude of freshwater inputs. Sampling sites are overlaid and coloured according to the six spatial clusters ($K=6$) derived from an augmented clustering space combining geographic coordinates and cumulative river influence.}
\label{fig:river_clusters}
\end{figure}

Sampling locations were partitioned into six coherent coastal clusters using a river-aware spatial clustering algorithm that combines geographic proximity with similarity in freshwater exposure, yielding distinct spatial regimes along the northern Portuguese coast (Figure~\ref{fig:river_clusters}). These clusters define the spatial blocking units in the cross-validation framework, ensuring that model evaluation reflects generalisation across physically distinct coastal environments. Full methodological details are provided in \ref{app:rofi_clustering}.
\subsection{Seasonal Features}
\label{subsec:seasonal_features}
\begin{figure}[H]
\centering
\includegraphics[width=0.95\textwidth]{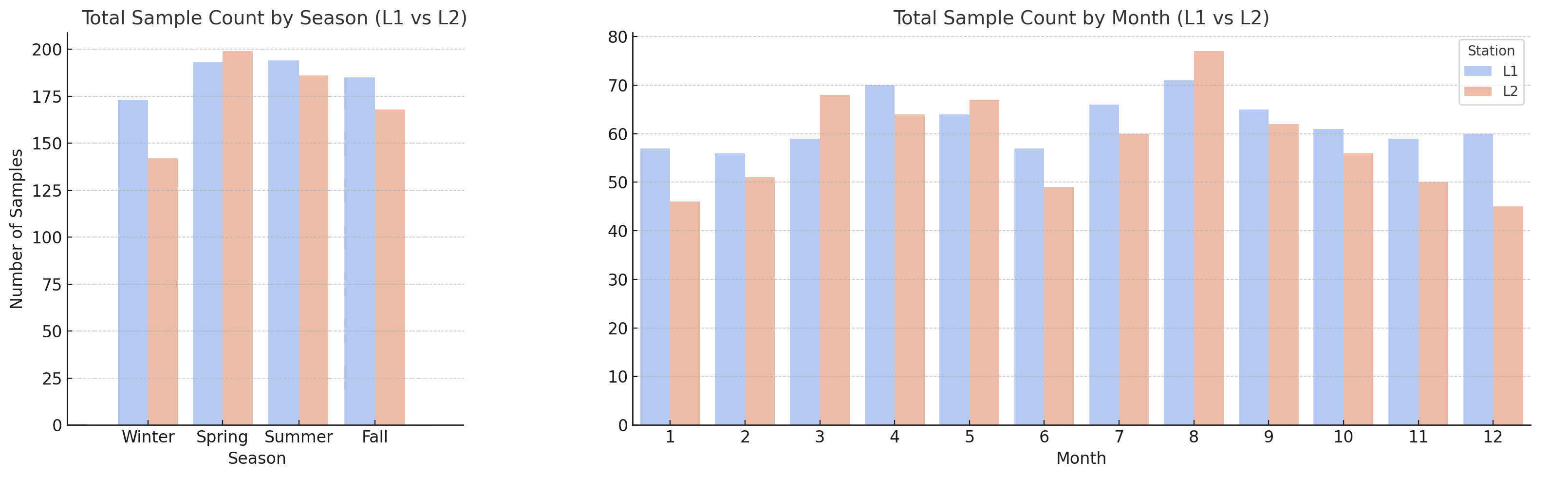}
\vspace{0.8em}
\caption{Seasonal and monthly distribution of sampling events (2013--2023) for the L1 and L2 production zones. Sampling intensity is highest during spring and summer, with pronounced monthly peaks in April and August, reflecting periods of elevated HAB risk.}
\label{fig:sampling_}
\end{figure}

Seasonality represents a dominant organising principle of PN dynamics along the Portuguese Atlantic coast and is therefore treated explicitly within the methodological framework. Both the ecological processes governing bloom development and the operational monitoring strategies implemented by IPMA exhibit strong seasonal structure, reflecting the coupling between atmospheric forcing, ocean circulation, and phytoplankton phenology. As described in Section~\ref{subsec:location_geo}, this seasonality is largely driven by the alternation between winter downwelling conditions and spring--summer upwelling regimes along the western Iberian margin. These seasonal cycles exert first-order control on PN occurrence and underpin both bloom timing and intensity.
The temporal distribution of \textit{in situ} sampling reflects this seasonal structure. Over the 2013--2023 period, monitoring effort in the northern production zones (L1 and L2) appears more concentrated during spring and summer. However, sampling is conducted on a regular (typically weekly) basis throughout the year as part of the national monitoring programme. Variations in sampling frequency primarily reflect operational constraints, such as adverse winter weather conditions, rather than deliberate seasonal targeting based on bloom risk. Sampling intensity increased markedly after 2016. This change is more likely associated with the progressive strengthening of monitoring protocols and regulatory compliance at the European level, including increased emphasis on phytoplankton monitoring and laboratory standardisation, rather than a direct response to short-term bloom variability (see Figure~\ref{fig:ts_spp_sst_ui}). While autumn and winter sampling effort is reduced, it is conducted regularly throughout the year; minor seasonal variations in sampling frequency may occur due to operational constraints such as adverse weather, but the schedule remains systematic, ensuring that off-season or anomalous bloom events are not excluded from the observational record. This adaptive strategy balances ecological relevance with operational constraints and introduces temporal heterogeneity that must be explicitly considered in predictive modelling. Seasonal aggregation of sampling events highlights a consistent bias toward spring and summer in both L1 and L2, corresponding to periods of enhanced upwelling intensity, nutrient supply, and phytoplankton growth (Figure~\ref{fig:sampling_}). Monthly distributions further refine this pattern, with distinct and persistent peaks in April and August. These months align with the typical onset of spring blooms and the recurrence of summer HAB events, respectively, as also indicated by PN cell concentrations and toxicity-related time series. The stability of these seasonal and monthly patterns across multiple years suggests a monitoring strategy that is strongly informed by long-term ecological experience and dominant bloom phenology.
To account for this pronounced seasonality in a manner suitable for machine-learning models, seasonal information was encoded explicitly as predictor variables rather than treated implicitly through time indexing. In addition to calendar descriptors (day, month, and year), the day of year (DOY) was transformed using sine and cosine functions to generate trigonometric seasonal harmonics:
\[
\textit{XDOY} = \sin\left(\frac{2\pi \cdot \mathrm{DOY}}{365}\right), \qquad
\textit{YDOY} = \cos\left(\frac{2\pi \cdot \mathrm{DOY}}{365}\right).
\]
This representation preserves the cyclical nature of the annual phenological cycle, avoids artificial discontinuities between December and January, and enables models to capture smooth seasonal transitions in bloom probability. These harmonic variables form the basis of the seasonal+spatial baseline model and are retained as covariates in all subsequent feature configurations, ensuring that seasonal structure is consistently accounted for when evaluating the incremental contribution of environmental and biological predictors.
\subsection{Environmental and Biological Features}
\label{subsec:env_bio_features}
Environmental and biological features constitute the core dynamic predictors used to characterise the physical forcing and ecological state associated with PN bloom development. All variables in this section are derived exclusively from satellite observations and are designed to represent processes that regulate nutrient supply, water-column structure, phytoplankton biomass, and community composition at spatial and temporal scales relevant to HAB dynamics along the Portuguese coast.
\paragraph{Environmental features (SST and upwelling index).}
Environmental predictors describe the large-scale physical conditions that modulate PN growth and bloom initiation. Two variables were selected: sea surface temperature (SST) and an empirically derived upwelling index (UI). SST captures seasonal warming and cooling, stratification strength, and air--sea coupling, all of which influence phytoplankton physiology and vertical nutrient exchange. SST data were obtained from the Copernicus Marine Environment Monitoring Service (CMEMS)\footnote{Copernicus Marine Service (2023), \href{https://marine.copernicus.eu/}{CMEMS portal}.} and collocated with \textit{in situ} observations using a consistent spatio-temporal matching procedure. Values were standardised to degrees Celsius (\(^\circ\)C) to ensure physical interpretability. The upwelling index (UI) provides a quantitative measure of wind-driven Ekman transport and coastal nutrient injection, a key component of the seasonal forcing along the western Iberian margin. UI was derived from surface wind stress following classical Ekman theory \parencite{ekman1905motion} and the coastal upwelling formulation of \textcite{bakun1973coastal}, with a default coastline angle of $-32^{\circ}$ for the western Iberian margin and zone-specific adjustments for L1 (Viana do Castelo) and L2 (Matosinhos) to account for documented variations in coastal geometry \parencite{fiuza1983upwelling}. Full details of the wind-stress, Coriolis and projection equations, as well as data retrieval, preprocessing and matching, are provided in \ref{app:cmems_pipeline}\footnote{\label{fn:cmems}Bnoussaad, A. (2024), ``CMEMS\_Data\_Analysis,'' \href{https://github.com/aymansvvd/CMEMS_Data_Analysis}{GitHub repository}.}. Together, SST and UI capture complementary aspects of physical forcing: SST reflects the thermal and stratification state of surface waters, while UI directly represents wind-driven nutrient supply. Both variables exhibit pronounced seasonal and interannual variability characteristic of the Iberian upwelling system and are expected to exert first-order control on PN bloom timing and intensity.
\paragraph{Biological features (chlorophyll-\emph{a} and plankton functional types).}
Biological predictors were selected to represent phytoplankton biomass and community structure beyond bulk chlorophyll alone. These variables were derived from the CMEMS North Atlantic Ocean Colour Bio-Geo-Chemical (BGC) product\footnote{Copernicus Marine Service (2025), ``Ocean Colour Product User Manual (Issue 6.0).''}, specifically the \texttt{bgc-plankton\_my\_l3-multi-1km\_P1D} dataset, which provides daily Level-3 estimates of chlorophyll-\textit{a} (CHL) concentration and multiple plankton functional types (PFTs) at 1~km spatial resolution. Biological predictors were collocated with \textit{in situ} sampling dates using a consistent spatio-temporal matching procedure. Extracted variables include chlorophyll-\textit{a} (CHL) and the full suite of available plankton functional types (PFTs) (Table~\ref{tab:dataset_summary}). Detailed preprocessing steps and quality-control procedures are described in \ref{app:cmems_pipeline}.
CHL provides a proxy for total phytoplankton biomass and overall productivity, while PFTs offer additional information on community composition relevant to PN ecology. These variables reflect phytoplankton succession across upwelling, stratification, and nutrient-depletion phases and enable discrimination between bloom-favourable assemblages and background communities \parencite{brewin2017PFTReview}. In addition to the central estimates, uncertainty fields associated with CHL and PFT retrievals were retained as auxiliary descriptors of bio-optical retrieval confidence in optically complex coastal waters. Although our framework leverages satellite ocean-colour products, including Chl-\textit{a} and phytoplankton functional types derived from Sentinel-3 OLCI via the CMEMS OCEANCOLOUR dataset \parencite{CMEMS_OCEANCOLOUR}, these variables remain empirical estimates obtained through bio-optical inversion. Their accuracy and ecological interpretability can therefore vary in optically complex coastal environments with strong riverine influence, suspended sediments, and diverse phytoplankton communities \parencite{IOCCG2014, melin2015, brewin2017PFTReview}. Consequently, while such products offer an improved representation of phytoplankton community structure relative to raw reflectance bands, their use in predictive models must explicitly account for underlying bio-optical uncertainty.
\paragraph{Temporal context and lagged features.}
To capture short-term environmental memory and delayed biological responses, lagged versions of the principal predictors were constructed. Environmental lags include SST and UI values from preceding days, while biological lags include CHL and PFT values from previous observations. Lag depths from 0 to 45 days were evaluated, reflecting ecologically plausible response times associated with nutrient enrichment, phytoplankton growth, and bloom development. All lagged variables were generated relative to the \textit{in situ} sampling date and appended as additional predictors, ensuring that only information available prior to or at the sampling time is used in model training. Implementation details of the data-processing pipeline are provided in \ref{app:cmems_pipeline}.
\subsection{Target Features}
\label{subsec:target_features}
The supervised targets used in this study are derived exclusively from long-term \textit{in situ} monitoring records provided by IPMA\footnote{Instituto Português do Mar e da Atmosfera (IPMA), \href{https://www.ipma.pt/pt/index.html}{IPMA website}. Accessed 2026.}. These records include PN cell concentrations (cells\,L$^{-1}$), as well as categorical information on DA-related harvesting bans. Target construction was guided by operational relevance, data availability, and the need to ensure physically realistic evaluation under strong class imbalance. PN cell concentration constitutes the primary biological observation underlying target definition. Across the L1--L2 modelling dataset (2013--2023), PN values span more than six orders of magnitude, ranging from zero to $3.63\times10^{6}$ cells\,L$^{-1}$ (Figure~\ref{fig:ts_spp_sst_ui}). The distribution is strongly right-skewed, reflecting episodic bloom events embedded within long periods of low background abundance. This extreme variability motivates a categorical formulation of bloom occurrence rather than direct regression on raw concentrations.

Harmful algal bloom (HAB) occurrence is defined as a binary classification target derived from PN cell concentrations using a fixed concentration threshold. Specifically, observations with PN $> 10{,}000$ cells\,L$^{-1}$ are classified as bloom events (\textit{HAB} = 1), while all remaining observations are classified as non-events (\textit{HAB} = 0). This threshold is consistent with commonly used operational benchmarks for elevated-risk \textit{Pseudo-nitzschia} conditions reported in the literature \parencite{lane2009logistic}. Compared to percentile-based definitions, the fixed threshold provides a more interpretable and transferable criterion aligned with monitoring practice. The resulting HAB class remains imbalanced, reflecting the episodic nature of bloom events in the observational record. By formulating HAB occurrence as the sole supervised target, the modelling framework is explicitly designed to emulate a remote-only early-warning system for PN blooms. All target labels are derived exclusively from \textit{in situ} observations and are never used as predictors, ensuring a strict separation between explanatory variables and response. This formulation allows model skill to be interpreted directly in an operational context, where the objective is to anticipate periods of elevated bloom risk that may warrant intensified monitoring, targeted sampling, or precautionary management actions.
\subsection{Dataset}
\begin{figure}[H]
\centering
\includegraphics[width=1.1\textwidth]{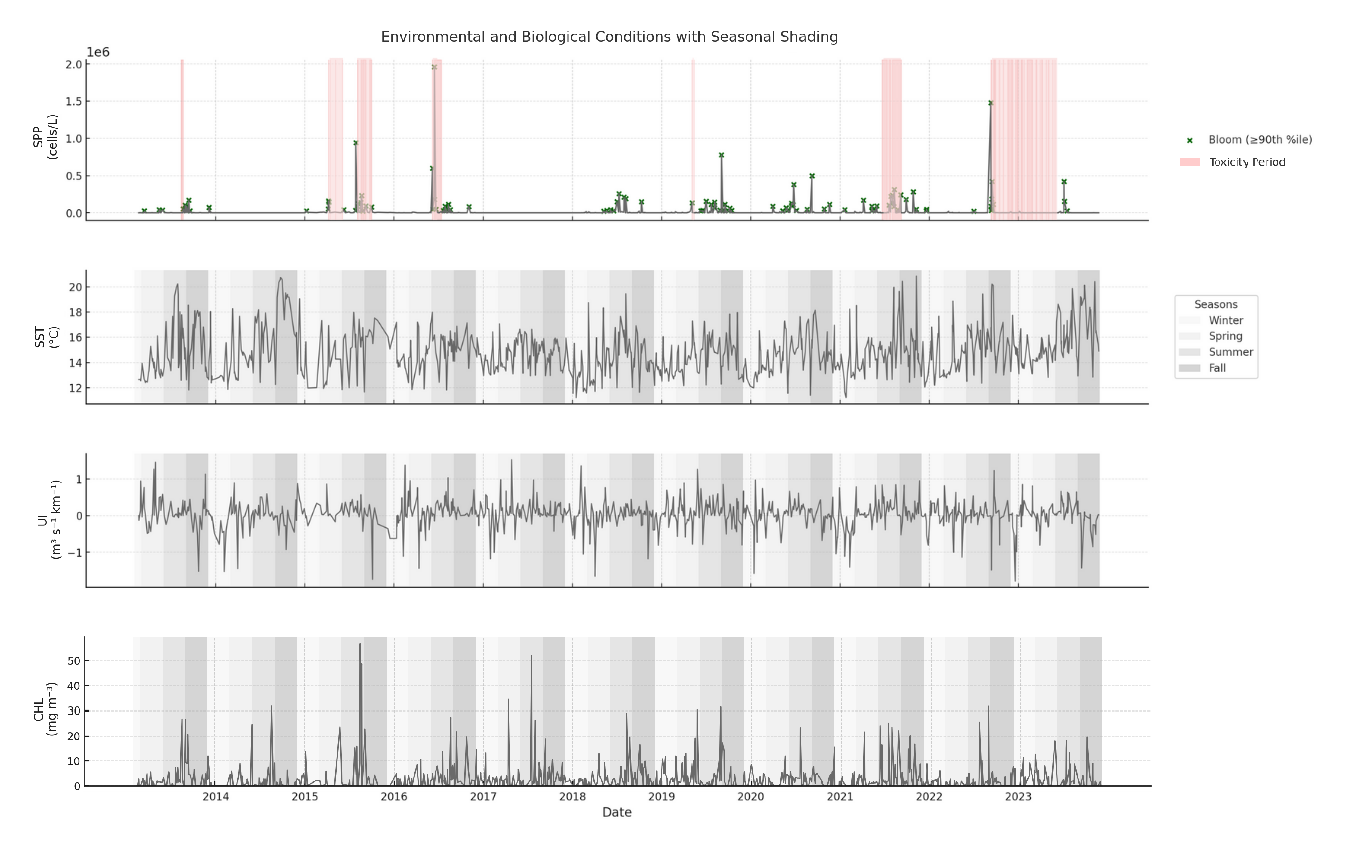}
\vspace{0.9em}
\caption{Time series (2013--2023) of (top) PN cell concentrations, with toxicity-related harvesting ban periods highlighted in red and identified HAB peaks marked in green; (middle) Sea Surface Temperature (SST, $^\circ$C) and Upwelling Index (UI, m$^{3}$\,s$^{-1}$\,km$^{-1}$); and (bottom) Chl-\textit{a} concentration (CHL, mg\,m$^{-3}$). Grey shading indicates seasonal intervals.}
\label{fig:ts_spp_sst_ui}
\end{figure}

All predictors and target variables were harmonised into a single spatio-temporally consistent matchup dataset that forms the basis for feature engineering, temporal lag analysis, and machine-learning model development. This section summarises the data integration strategy and the main characteristics of the compiled dataset, while subsequent subsections detail the individual methodological components.
\paragraph{Regional multi-zone context (L1--L9).}
An initial descriptive statistical analysis was conducted on the full multi-zone dataset encompassing production zones L1 through L9 (5,882 observations), providing a system-wide characterisation of environmental variability and biological productivity along the Portuguese coast (Figure~\ref{fig:fig1}). Descriptive statistics were computed independently for each zone to characterise central tendencies, dispersion, and overall variability across the study region. Sample sizes varied among zones, reflecting differences in monitoring accessibility, production activity, and long-term data availability. Zones such as L1, L2, L7a, and L7c2 exhibited the largest number of observations (exceeding 500 records each), whereas zones such as L4 and L7b had comparatively fewer records due to zone-specific operational constraints, including offshore sampling conditions and historically irregular or discontinued production activity. Mean primary production (PN) displayed pronounced spatial heterogeneity across zones, with the highest average values observed in northern (L1, L2) and offshore-influenced regions (notably L7a and L7c2, where mean PN exceeded 29{,}000~cells\,L$^{-1}$), and substantially lower means in zones such as L4 and L7b, where average production remained below 10{,}000~units. Variability in PN was also zone-dependent, with standard deviations frequently comparable in magnitude to the corresponding means, indicating highly dynamic production regimes. Collectively, these results demonstrate strong spatial structuring in biological productivity across the L1--L9 production system and motivate the subsequent focus on zone-resolved, matched datasets.
\paragraph{L1--L2 modelling dataset.}
Building on the regional context above, predictive modelling focuses on the northern production zones Litoral de Viana (L1) and Litoral de Matosinhos (L2), which together define the core study region. The modelling dataset spans 2013--2023, with each row representing one IPMA sampling event augmented with matched satellite-derived predictors, associated uncertainty estimates, and lagged histories (0--45 days). Predictor groups include environmental variables (SST and UI), biological variables (CHL and PFTs), geographical coordinates, and temporal descriptors capturing seasonal structure; target variables derived from monitoring records include PN cell concentration and HAB occurrence. Satellite predictors were derived from two complementary CMEMS data streams. The environmental stream comprises SST and the empirically derived upwelling index (UI), representing large-scale physical forcing linked to seasonal warming, wind-driven upwelling, and relaxation events. Biological predictors were derived from the CMEMS BGC product described in Section~\ref{subsec:env_bio_features}. Together, these variables capture both bulk phytoplankton biomass and shifts in community structure relevant to \textit{Pseudo-nitzschia} ecology. PN cell concentrations (cells\,L$^{-1}$) exhibit extreme variability (Figure~\ref{fig:ts_spp_sst_ui}), ranging from 0 to $3.63\times10^{6}$ cells\,L$^{-1}$. The distribution is strongly right-skewed, reflecting episodic bloom events embedded within prolonged low-abundance periods. HAB occurrences are comparatively rare, consistent with the strong class imbalance typical of operational HAB forecasting. Environmental predictors show pronounced seasonal dynamics characteristic of the Iberian upwelling system. SST ranges from 10.94 to 20.90\,$^\circ$C (mean = 14.74\,$^\circ$C), while UI spans $-2.24$ to 2.29\,m$^{3}$\,s$^{-1}$\,km$^{-1}$ (mean = 0.034). Biological variables also display strong variability: CHL ranges from 0.13 to 97.11\,mg\,m$^{-3}$ (mean = 22.07\,mg\,m$^{-3}$), with diatom-associated biomass (DIATO) reaching 40.66\,mg\,m$^{-3}$ under bloom-favourable conditions. Other PFTs, including dinoflagellates (DINO), haptophytes (HAPTO), and pico- and nanophytoplankton groups (PICO, NANO), exhibit distinct seasonal cycles, reflecting phytoplankton succession across periods of upwelling, stratification, and nutrient depletion.
A statistical summary of all environmental and biological variables included in the matchup dataset is presented in Table~\ref{tab:dataset_summary}. For the binary variable (HAB occurrence), the mean represents event frequency (about 14\% for HAB events). These descriptive statistics provide quantitative context for the magnitude, variability, and imbalance inherent in the data and inform subsequent feature transformation, lag selection, and model design. Collectively, this integrated dataset links the environmental drivers and biological state variables to our bloom observations within a single analytical framework, forming the empirical basis for all subsequent modelling steps.
\begin{table}[H]
\centering
\small
\caption{Descriptive statistics of the L1--L2 modelling dataset (2013--2023), grouped by data origin and variable type. Remote-sensed predictors include satellite-derived environmental and biological variables and their temporal lags (0--45 days). Units: SST in $^\circ$C; UI in m$^{3}$\,s$^{-1}$\,km$^{-1}$; CHL and PFTs in mg\,m$^{-3}$; PN in cells\,L$^{-1}$.}
\label{tab:dataset_summary}
\begin{tabular}{llcccc}
\hline
\textbf{Data source} & \textbf{Variable (grouped)} & \textbf{Min} & \textbf{Max} & \textbf{Mean} & \textbf{Std. Dev.} \\
\hline
\multicolumn{6}{l}{\textit{Target variable}} \\
PN & \textit{Pseudo-nitzschia} cell concentration & 0 & $3.63\times10^{6}$ & $1.85\times10^{4}$ & $1.25\times10^{5}$ \\
HAB & Bloom occurrence (binary) & 0 & 1 & 0.14 & 0.34 \\
\hline
\multicolumn{6}{l}{\textit{Remote-sensed environmental predictors (lags 0--45)}} \\
SST & Sea surface temperature & 10.94 & 20.90 & 14.74 & 1.76 \\
UI & Upwelling index & $-2.24$ & 2.29 & 0.034 & 0.39 \\
\hline
\multicolumn{6}{l}{\textit{Remote-sensed biological predictors (lags 0--45)}} \\
CHL & Chlorophyll-\textit{a} & 0.13 & 97.11 & 22.07 & 23.06 \\
DIATO & Diatoms & 0.001 & 40.66 & 0.27 & 0.93 \\
DINO & Dinoflagellates & 0.001 & 8.84 & 0.17 & 0.26 \\
HAPTO & Haptophytes & 0.005 & 2.25 & 0.10 & 0.17 \\
GREEN & Green algae & 0.011 & 10.00 & 2.35 & 2.34 \\
PROKAR & Prokaryotes & 0.005 & 0.16 & 0.028 & 0.013 \\
PROCHLO & Prochlorococcus & 0.001 & 1.00 & 0.57 & 0.47 \\
MICRO & Microplankton & 0.005 & 49.50 & 0.44 & 1.15 \\
NANO & Nanoplankton & 0.005 & 2.25 & 0.10 & 0.17 \\
PICO & Picoplankton & 0.019 & 10.00 & 2.37 & 2.33 \\
\hline
\multicolumn{6}{l}{\textit{Remote-sensed biological uncertainty estimates (lags 0--45)}} \\
CHL uncertainty & Chlorophyll-\textit{a} retrieval uncertainty & 89.18 & 327.67 & 212.41 & 44.45 \\
PFT uncertainty & All phytoplankton functional types & 38.32 & 327.67 & 231.9 & 53.8 \\
\hline
\multicolumn{6}{l}{\textit{In situ temporal variables}} \\
Day & Day of month & 1 & 31 & 15.5 & 8.7 \\
Month & Month of year & 1 & 12 & 6.5 & 3.3 \\
Year & Calendar year & 2013 & 2023 & 2018 & 2.86 \\
DayOfYear & Julian day & 2 & 364 & 182.2 & 101.7 \\
XDOY & Seasonal harmonic (cosine) & $-1.00$ & 1.00 & 0.001 & 0.72 \\
YDOY & Seasonal harmonic (sine) & $-1.00$ & 1.00 & $-0.044$ & 0.69 \\
\hline
\multicolumn{6}{l}{\textit{In situ geographical variables}} \\
LATITUDE & Latitude & 41.03 & 41.75 & 41.41 & 0.31 \\
LONGITUDE & Longitude & $-8.89$ & $-8.70$ & $-8.79$ & 0.08 \\
\hline
\end{tabular}
\end{table}
\subsection{Model Training and Machine-Learning Framework}
\label{subsec:model_training}
The harmonised L1--L2 matchup dataset was used to train supervised classifiers for HAB occurrence prediction under the spatio-temporal validation framework described below. All predictive experiments rely exclusively on remotely sensed environmental and biological variables, together with seasonal indicators and spatial coordinates, thereby emulating a remote-only operational early-warning setting.
\subsection{Spatio-temporal Cross-Validation and Data Partitions}
\label{subsec:spatio_temporalCV}
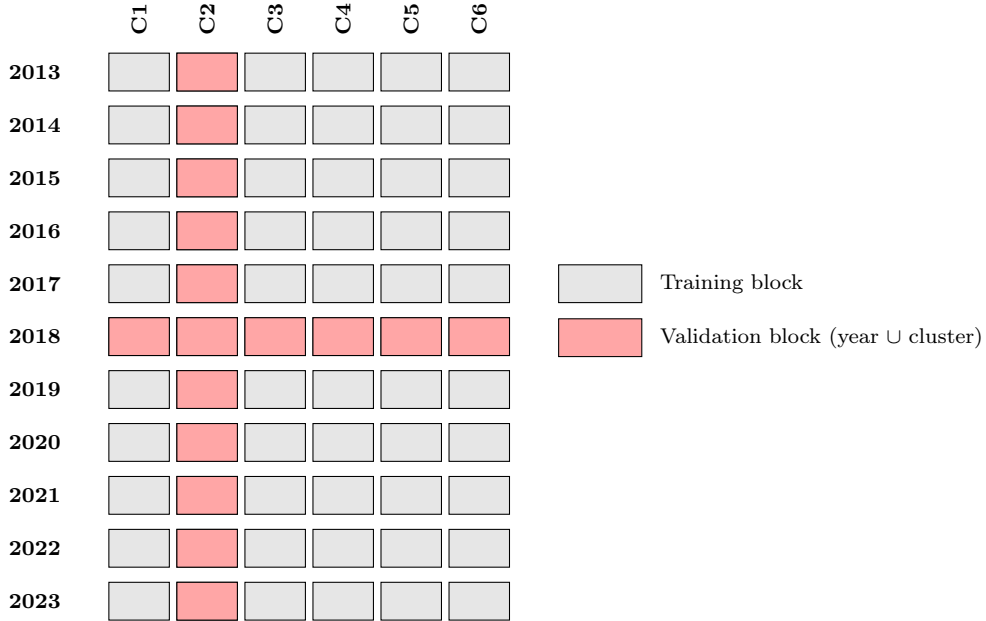
\begin{figure}[H]
\centering
\begin{tikzpicture}[
    cell/.style={draw, minimum width=8mm, minimum height=5mm, font=\scriptsize, align=center},
    train/.style={cell, fill=gray!20},
    test/.style={cell, fill=red!35},
    yearlabel/.style={font=\scriptsize},
    clusterlabel/.style={font=\scriptsize, rotate=90}
]
\def\years{2013,2014,2015,2016,2017,2018,2019,2020,2021,2022,2023}
\def\nyears{11}
\def\nclus{6}
\foreach \j in {1,...,\nclus} {
    \node[clusterlabel] at (0.9*\j,0) {\textbf{C\j}};
}
\foreach \i [count=\yi] in \years {
    \node[yearlabel, anchor=east] at (0, -0.7*\yi) {\textbf{\i}};
    \foreach \j in {1,...,\nclus} {
        \node[train] at (0.9*\j, -0.7*\yi) {};
    }
}
\foreach \j in {1,...,\nclus} {
    \node[test] at (0.9*\j, -0.7*6) {};
}
\foreach \i [count=\yi] in \years {
    \node[test] at (0.9*2, -0.7*\yi) {};
}
\begin{scope}[shift={(7, -3.5)}]
    \node[cell, fill=gray!20, minimum width=11mm] (l1) {};
    \node[right=1mm of l1, font=\scriptsize] {Training block};
    \node[cell, fill=red!35, minimum width=11mm, below=2mm of l1] (l2) {};
    \node[right=1mm of l2, font=\scriptsize] {Validation block (year $\cup$ cluster)};
\end{scope}
\end{tikzpicture}
\caption{Spatio-temporal cross-validation schematic. For each fold, all observations from one target year (entire row) and one spatial cluster (entire column) are simultaneously held out (red), while all remaining spatio-temporal blocks (grey) are used for training. This prevents leakage in both time and space and reflects realistic forecasting constraints.}
\label{fig:stcv}
\end{figure}

To obtain realistic generalisation estimates for HAB occurrence, all models were evaluated under a spatio-temporal cross-validation scheme that preserves both temporal dependence and coastal heterogeneity. Conventional random (i.i.d.) splits would artificially inflate predictive performance by allowing information leakage across nearby stations and adjacent sampling dates. Instead, the evaluation framework enforces strict separation in both time and space by withholding entire calendar years and spatial clusters simultaneously, as described in Section~\ref{subsec:location_geo}. Spatial structure was defined by river-aware $K$-means clustering of sampling locations, yielding 6 spatial clusters, while temporal structure was defined by calendar year over an 11-year record (2013--2023). This design produced 66 validation splits (11 years $\times$ 6 clusters). In each split, the validation set consisted of all observations from one selected calendar year together with all observations from one selected spatial cluster, such that any sample matching either condition was excluded from training. Environmental (SST, UI) and biological (CHL, PFT) predictors were supplemented with lagged features (up to 45 days) constructed relative to the sampling date. Importantly, spatio-temporal splitting was performed on the original timestamps and spatial cluster assignments, ensuring that no information from a held-out year or coastal segment could enter the training folds through lagged variables. As a consequence, reported performance metrics reflect the model's ability to generalise to previously unseen years and spatial regimes rather than its capacity to interpolate within known conditions. Figure~\ref{fig:stcv} illustrates the structure of the spatio-temporal cross-validation scheme.
\subsubsection{Machine-Learning Models and Training Procedure}
A suite of machine-learning classifiers was evaluated to predict HAB occurrence, spanning linear, margin-based, and non-linear ensemble approaches. The tested models include logistic regression (LR), support-vector machines (SVM), random forests (RF), extremely randomised trees (Extra Trees), and gradient-boosted decision trees (XGBoost). These models were applied consistently across the three predictor configurations defined earlier: (i) a seasonal--spatial baseline, (ii) a seasonal--spatial--environmental configuration, and (iii) a fully enriched configuration incorporating seasonal, spatial, environmental, and biological predictors. This model set was selected to balance interpretability and representational flexibility. Logistic regression provides a transparent linear benchmark, while SVM offers a margin-based classifier capable of non-linear decision boundaries depending on kernel specification. Tree-based ensemble methods are particularly well suited to the present problem, as they can capture non-linear relationships, threshold effects, and interactions among predictors with minimal parametric assumptions. Such properties are advantageous given the complex coupling between physical forcing, biological state, and bloom dynamics. All models were trained exclusively on training folds defined by the spatio-temporal cross-validation scheme.
To ensure reproducibility and clarify the source of the reported performance, we provide the final hyperparameter configurations of the best-performing models for each feature setting. Hyperparameters were selected through our cross-validation within the training folds, and the reported configurations correspond to those achieving the highest mean ROC--AUC, a performance metric described in detail in the following subsection. In the environmental-only configuration, the Random Forest classifier achieved the highest performance. The final configuration used 400 trees (\texttt{n\_estimators = 400}), a maximum tree depth of 8 (\texttt{max\_depth = 8}), and a minimum of 3 samples per terminal leaf (\texttt{min\_samples\_leaf = 3}). This combination provided stable performance across folds while limiting excessive tree growth and overly specific partitioning of rare HAB events. Because HAB events are ecologically episodic yet operationally high-consequence, the positive class represents only a small fraction of the observational record. To ensure that these bloom episodes remained appropriately weighted during model training, we applied balanced class weights (\texttt{class\_weight = balanced}), preventing the classifier from being dominated by the more frequent non-bloom background conditions.

Likewise, for the full feature configuration (all features), the Extremely Randomised Trees (Extra Trees) classifier yielded the best overall performance with 200 trees and a minimum of 2 samples per leaf. Compared to Random Forest, Extra Trees introduces more randomness during its decision-making process, which tends to improve performance when many correlated environmental predictors are present. Class imbalance was handled using balanced class weights, and a fixed random seed of 42 was used throughout.
\subsubsection{Performance Metrics and Feature Importance}
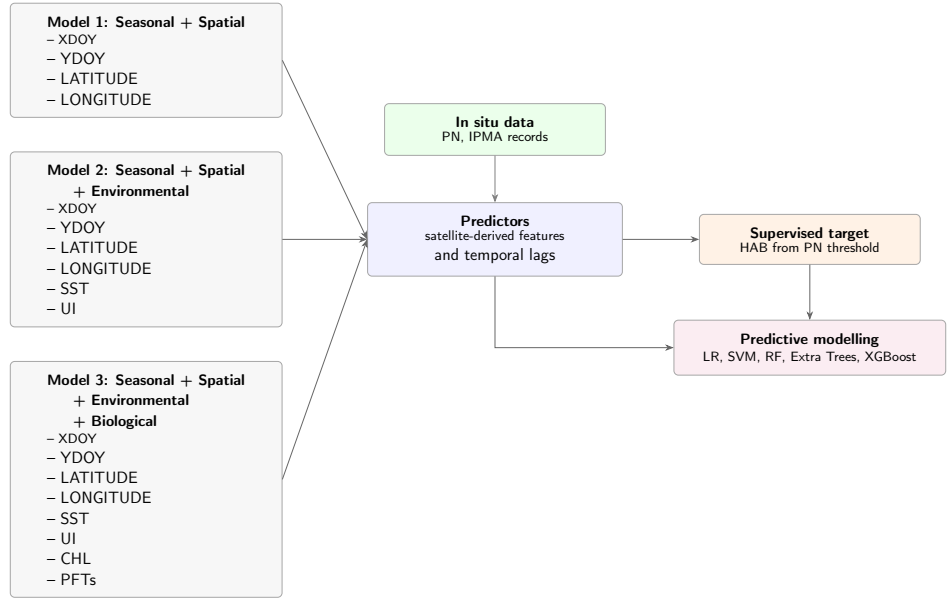
\begin{figure}[H]
\centering
\resizebox{0.75\linewidth}{!}{%
\sffamily
\begin{tikzpicture}[
  node distance=8mm and 18mm,
  >=Stealth
]
\node[
  draw=black!35, rounded corners=3pt, line width=0.4pt,
  fill=gray!6, align=left, inner xsep=5mm, inner ysep=3mm,
  minimum width=64mm
] (m1) {%
  \textbf{\small Model 1: Seasonal + Spatial}\\[-0.2em]
  \footnotesize
  -- XDOY\\
  -- YDOY\\
  -- LATITUDE\\
  -- LONGITUDE
};
\node[
  draw=black!35, rounded corners=3pt, line width=0.4pt,
  fill=gray!6, align=left, inner xsep=5mm, inner ysep=3mm,
  minimum width=64mm,
  below=of m1
] (m2) {%
  \textbf{\small Model 2: Seasonal + Spatial}\\
  \hspace{1.6em}\textbf{\small + Environmental}\\[-0.2em]
  \footnotesize
  -- XDOY\\
  -- YDOY\\
  -- LATITUDE\\
  -- LONGITUDE\\
  -- SST\\
  -- UI
};
\node[
  draw=black!35, rounded corners=3pt, line width=0.4pt,
  fill=gray!6, align=left, inner xsep=5mm, inner ysep=3mm,
  minimum width=64mm,
  below=of m2
] (m3) {%
  \textbf{\small Model 3: Seasonal + Spatial}\\
  \hspace{1.6em}\textbf{\small + Environmental}\\
  \hspace{1.6em}\textbf{\small + Biological}\\[-0.2em]
  \footnotesize
  -- XDOY\\
  -- YDOY\\
  -- LATITUDE\\
  -- LONGITUDE\\
  -- SST\\
  -- UI\\
  -- CHL\\
  -- PFTs
};
\node[
  draw=black!35, rounded corners=3pt, line width=0.4pt,
  fill=blue!6, align=center, inner xsep=6mm, inner ysep=3mm,
  minimum width=60mm,
  right=20mm of m2
] (pred) {%
  \textbf{\small Predictors}\\[-0.4em]
  \footnotesize satellite-derived features\\
  and temporal lags
};
\node[
  draw=black!35, rounded corners=3pt, line width=0.4pt,
  fill=green!8, align=center, inner xsep=5mm, inner ysep=3mm,
  minimum width=52mm,
  above=11mm of pred
] (insitu) {%
  \textbf{\small In situ data}\\[-0.4em]
  \footnotesize PN, IPMA records
};
\node[
  draw=black!35, rounded corners=3pt, line width=0.4pt,
  fill=orange!10, align=center, inner xsep=5mm, inner ysep=3mm,
  minimum width=52mm,
  right=18mm of pred
] (target) {%
  \textbf{\small Supervised target}\\[-0.4em]
  \footnotesize HAB from PN threshold
};
\node[
  draw=black!35, rounded corners=3pt, line width=0.4pt,
  fill=purple!7, align=center, inner xsep=6mm, inner ysep=3mm,
  minimum width=64mm,
  below=13mm of target
] (model) {%
  \textbf{\small Predictive modelling}\\[-0.2em]
  \footnotesize LR, SVM, RF, Extra Trees, XGBoost
};
\draw[-Stealth, line width=0.6pt, draw=black!60] (m1.east) -- (pred.west);
\draw[-Stealth, line width=0.6pt, draw=black!60] (m2.east) -- (pred.west);
\draw[-Stealth, line width=0.6pt, draw=black!60] (m3.east) -- (pred.west);
\draw[-Stealth, line width=0.55pt, draw=black!50] (insitu.south) -- (pred.north);
\draw[-Stealth, line width=0.6pt, draw=black!60] (pred.east) -- (target.west);
\draw[-Stealth, line width=0.6pt, draw=black!60] (target.south) -- (model.north);
\draw[-Stealth, line width=0.6pt, draw=black!60] (pred.south) |- (model.west);
\end{tikzpicture}%
}
\caption{Schematic of predictor configurations and modelling flow. Three progressively enriched predictor sets are evaluated: (Model~1) seasonal and spatial baseline covariates, (Model~2) seasonal and spatial covariates plus environmental predictors, and (Model~3) seasonal and spatial covariates plus environmental and biological predictors. Predictors are derived from satellite products (including lagged histories), while the supervised HAB target is constructed from \textit{in situ} PN concentrations.}
\label{fig:dataset_workflow}
\end{figure}

Model performance was assessed using a complementary set of threshold-dependent and threshold-independent metrics. These include Accuracy, Precision, Recall, and F1-score, as well as the area under the receiver operating characteristic curve (ROC--AUC). ROC--AUC summarises a model's ability to rank HAB versus non-HAB observations across all possible decision thresholds by integrating the receiver operating characteristic curve. An AUC of 0.5 corresponds to random discrimination, whereas a value of 1.0 indicates perfect ranking. Because ROC--AUC is threshold-independent and relatively insensitive to class imbalance, it provides a robust scalar metric for comparing models under cross-validation. The HAB target exhibits strong class imbalance (approximately 14\% positive events), which can increase variance in fold-wise performance metrics and amplify sensitivity to threshold-dependent measures. The use of ROC--AUC mitigates this issue by providing a threshold-independent assessment of ranking ability, while the spatio-temporal cross-validation framework ensures that each fold contains structurally distinct event distributions. Consequently, the reported standard deviations reflect variability in model performance across heterogeneous environmental regimes rather than instability due to class imbalance alone.
To interpret model behaviour and identify the primary environmental drivers of HAB occurrence, feature importance analyses were conducted for the tree-based ensemble models. For random forests and extremely randomised trees, feature importance was first quantified using the mean decrease in impurity (MDI), which identifies which environmental and biological factors; such as sea surface temperature, upwelling intensity or chlorophyll concentration, are most effective at distinguishing bloom from non-bloom conditions throughout the model's decision structure \parencite{breiman2001random}. While impurity-based importance provides a computationally efficient summary of how frequently and effectively variables are used in tree construction, it is known to exhibit biases in the presence of correlated predictors or variables with differing scales and cardinalities. To obtain a more robust and model-agnostic assessment, impurity-based rankings were complemented with permutation importance analysis \parencite{breiman2001random,strobl2008conditional}. In this approach, the values of each predictor are randomly permuted in the validation data, and the resulting decrease in predictive performance is measured. Features whose permutation leads to a large degradation in model skill are interpreted as having high predictive relevance. Because permutation importance evaluates the contribution of predictors with respect to out-of-sample performance, it provides a more reliable indication of feature relevance under correlated and non-linear settings.
For gradient-boosted trees, importance scores based on split gain were examined in parallel, reflecting how much each environmental and biological variable directly improved the model's ability to reduce prediction errors at each successive step of the boosting process. Feature importance analyses were performed within the cross-validation framework to ensure that rankings reflect generalisable predictive structure rather than artefacts of individual training folds. Collectively, these complementary importance measures allow identification of robust ecological signals, such as seasonal timing, physical forcing, and biological context, while supporting mechanistic interpretation of model outputs in an operational forecasting setting.
To synthesise the methodological choices described above and clarify the structure of the predictive experiments, Figure~\ref{fig:dataset_workflow} provides a schematic overview of the modelling framework. The diagram illustrates the progressive construction of predictor sets, ranging from a seasonal+spatial baseline to increasingly enriched configurations that incorporate satellite-derived environmental and biological information. All predictor sets are derived exclusively from remotely sensed data, including temporally lagged conditions, while the supervised HAB target is defined independently from \textit{in situ} PN cell concentrations. This design ensures a clear separation between predictors and labels and reflects an operational forecasting context in which bloom risk must be inferred from satellite observations alone.

\section{Results}
\subsection{Discrimination performance and ROC--AUC}
\begin{figure}[H]
\centering
\includegraphics[width=0.55\textwidth]{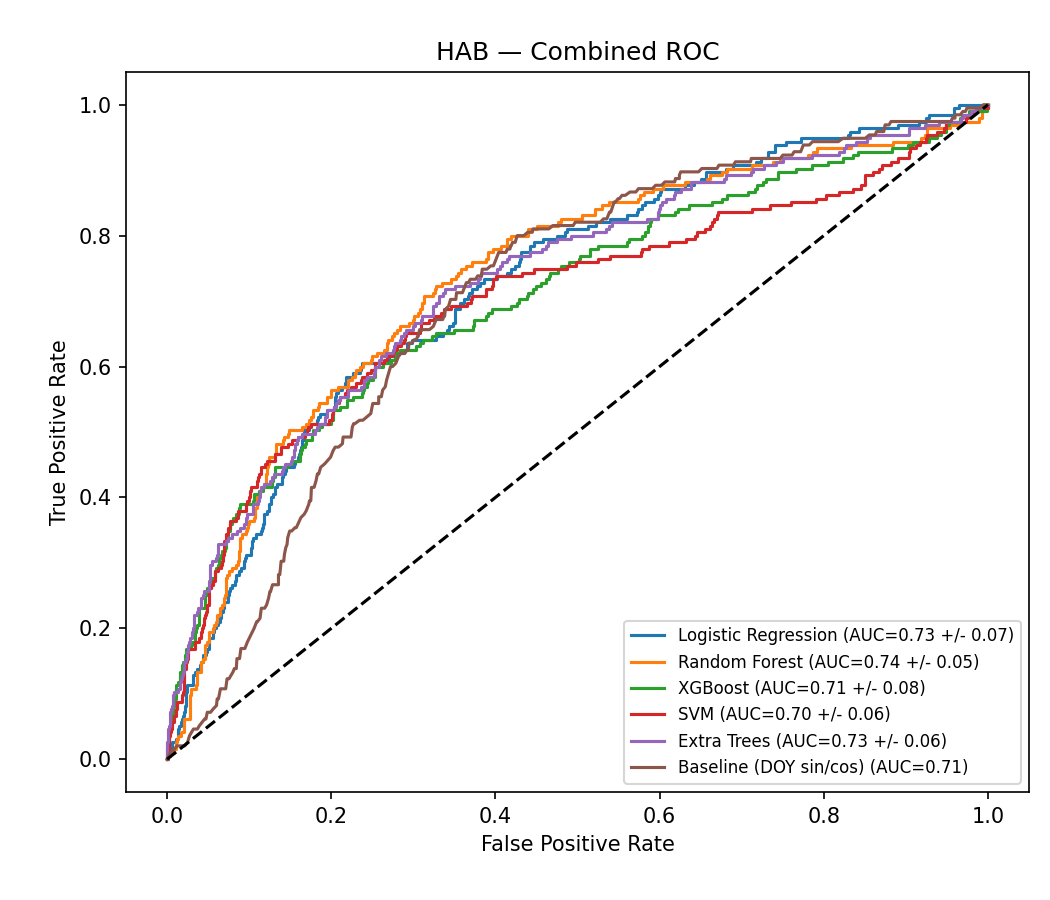}
\caption{ROC curves for the HAB target using Environmental-only features. Ensemble tree models (Random Forest, Extra Trees, XGBoost) generally outperform linear and kernel baselines across most of the ROC domain.}
\label{fig:roc_hab_env}
\end{figure}

Model predictive accuracy was evaluated using the ROC--AUC metric, computed from out-of-fold predictions generated by the river-aware spatio-temporal cross-validation scheme. ROC--AUC provides a threshold-independent measure of a model's ability to distinguish bloom events from non-event conditions and is therefore well suited to the imbalanced nature of HAB occurrence. Reported ROC--AUC values are expressed as mean $\pm$ standard deviation across different test scenarios, where each fold corresponds to a distinct combination of withheld year and spatial cluster. Because validation is performed on fully unseen spatio-temporal domains, this variability reflects genuine sensitivity to out-of-distribution conditions rather than statistical noise or resampling artefacts. Thus, the model was repeatedly challenged to forecast bloom occurrence in geographic areas and specific years that were entirely absent from its training data, making this a stringent test of genuine predictive capability rather than pattern memorisation. Table~\ref{tab:results_auc} summarises ROC--AUC values for HAB prediction across two feature configurations: an Environmental-only configuration (seasonality, coordinates, SST and UI, including lagged values), and an Environmental + Biological configuration that additionally incorporates Chl-\textit{a} and plankton functional types (CHL and PFTs). Lagged predictors are identical across configurations (45 days); differences arise solely from the inclusion of biological information. The seasonal+spatial baseline (logistic regression using only XDOY/YDOY and coordinates) is shown as a reference model.
Across all models, the ability to distinguish HAB events ranged from approximately $0.66$ to $0.77$; these values, while moderate, represent meaningful predictive skill considering that the models rely exclusively on satellite-derived observations, with no local \textit{in situ} measurements of any kind used as predictors. Ensemble tree methods consistently outperformed linear and kernel-based classifiers, and incorporating biological predictors alongside environmental forcing further improved discrimination, demonstrating that community-level satellite information adds value beyond physical forcing alone. The best overall performance is achieved by the Extra Trees classifier with Environmental + Biological features (ROC--AUC = $0.77 \pm 0.06$). These performance estimates reflect how the model performs under truly unseen conditions, predicting bloom occurrence in years and coastal areas that were entirely absent from training, rather than interpolating within familiar spatio-temporal patterns. An AUC of $0.77 \pm 0.06$ obtained under these conditions indicates that the model retains meaningful discriminative ability when confronted with environmental regimes it has never encountered, which is the relevant test for any system intended for operational forecasting. This contrasts with substantially higher AUC values (e.g.\ $>0.90$) often reported under random or weakly structured cross-validation, which can be driven by spatial and temporal autocorrelation and therefore reflect partial memorisation of recurring patterns rather than transferable predictive skill. By using a strict validation protocol, our results provide a conservative but realistic estimate of how well this system would perform in an operational setting predicting PN blooms.
Moreover, preprocessing was fitted within each training split only, and predictions were generated exclusively for held-out samples. Out-of-fold predicted probabilities from all splits were then pooled to calculate an overall ROC--AUC, providing a cross-validated estimate of discrimination based entirely on unseen data. To quantify uncertainty, a fold-specific ROC--AUC was also computed for each of the 66 validation splits, and the mean and standard deviation of these fold-wise AUC values were used to summarise performance variability across spatio-temporal partitions. Accordingly, the reported AUC reflects pooled out-of-fold performance, whereas the accompanying uncertainty captures variation across the year-by-cluster validation splits.
\begin{table}[H]
\centering
\caption{ROC--AUC performance for HAB prediction (mean $\pm$ standard deviation across 66 spatio-temporal cross-validation folds).}
\label{tab:results_auc}
\begin{tabular}{lcc}
\hline
\textbf{Model} & \textbf{Env.} & \textbf{Env.\,+\,Bio.} \\
\hline
Logistic Regression & $0.73 \pm 0.07$ & $0.67 \pm 0.04$ \\
Random Forest       & $\mathbf{0.74 \pm 0.05}$ & $0.76 \pm 0.05$ \\
XGBoost             & $0.71 \pm 0.08$ & $0.75 \pm 0.06$ \\
SVM                 & $0.70 \pm 0.06$ & $0.66 \pm 0.07$ \\
Extra Trees         & $0.73 \pm 0.06$ & $\mathbf{0.77 \pm 0.06}$ \\
Baseline (LR)       & $0.71$          & $0.71$ \\
\hline
\end{tabular}
\end{table}
\paragraph{Environmental-only features.}
Using only seasonal/spatial indicators together with SST and UI (including lagged values), HAB discrimination ranges between $0.70 \pm 0.06$ and $0.74 \pm 0.05$ (Table~\ref{tab:results_auc}). The Random Forest model achieves the highest ROC--AUC in this configuration ($0.74 \pm 0.05$), followed closely by Logistic Regression and Extra Trees. The seasonal/spatial baseline (LR) achieves a ROC--AUC of 0.71, indicating that environmental forcing provides a measurable improvement over climatological structure alone. The corresponding ROC curves are shown in Figure~\ref{fig:roc_hab_env}.

\begin{figure}[H]
\centering
\includegraphics[width=0.55\textwidth]{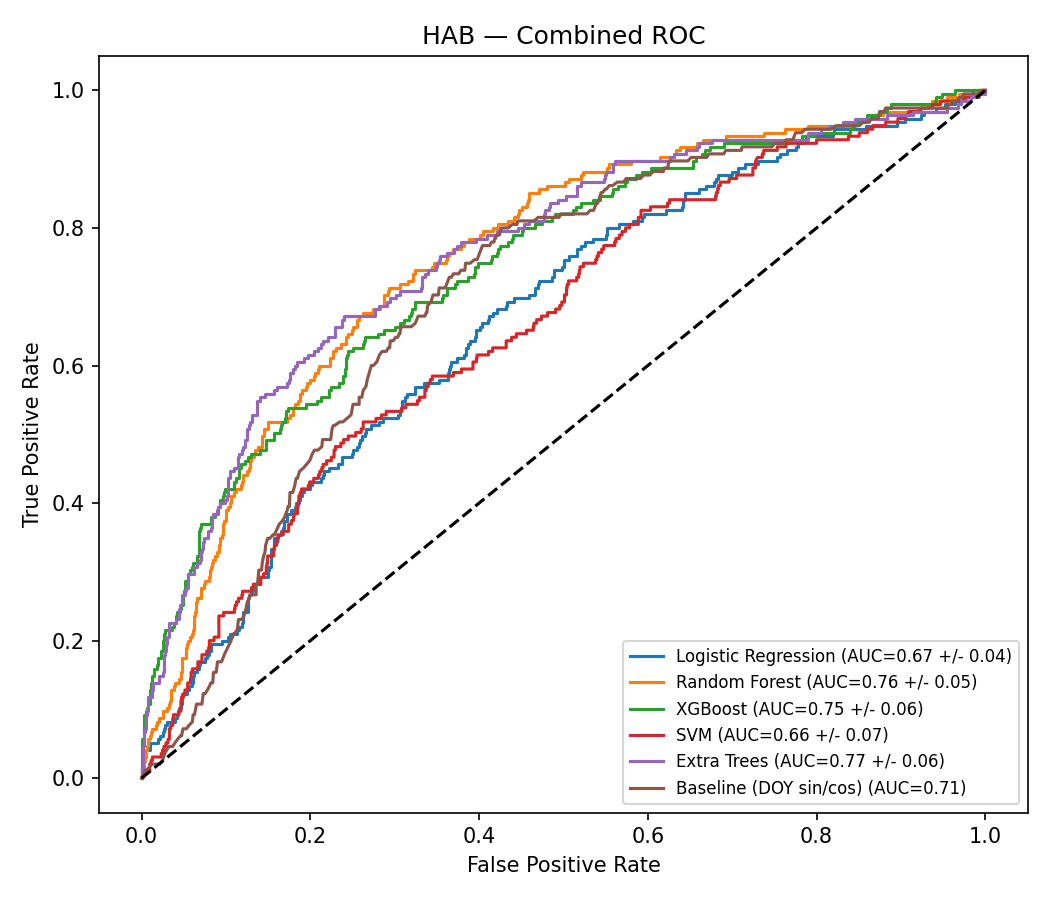}
\caption{ROC curves for the HAB target using Environmental + Biological features. Incorporating CHL and PFTs improves discrimination for ensemble models, with Extra Trees achieving the strongest overall performance.}
\label{fig:roc_hab_bio}
\end{figure}

\paragraph{Environmental + Biological features.}
Incorporating biological predictors (CHL and PFTs) yields a clear improvement in HAB discrimination for the ensemble models. Extra Trees attains the highest overall performance (ROC--AUC = $0.77 \pm 0.06$), with Random Forest ($0.76 \pm 0.05$) and XGBoost ($0.75 \pm 0.06$) closely following. The corresponding ROC curves (Figure~\ref{fig:roc_hab_bio}) exhibit steeper initial slopes, indicating stronger early separation between bloom and non-bloom conditions. In contrast, Logistic Regression and SVM exhibit reduced performance when biological predictors are included, suggesting that the additional information is expressed through non-linear interactions that are not effectively captured by linear or margin-based models.

Overall, the results demonstrate that environmental forcing and seasonal/spatial structure alone provide a robust baseline for HAB prediction, while the inclusion of satellite-derived biological information significantly enhances discrimination when combined with flexible, non-linear learning algorithms. These findings support the use of biologically enriched feature sets for operational HAB early-warning systems along the Portuguese coast.
\subsection{Feature importance and operational performance for HAB prediction}
\begin{figure}[H]
\centering
\includegraphics[width=0.95\textwidth]{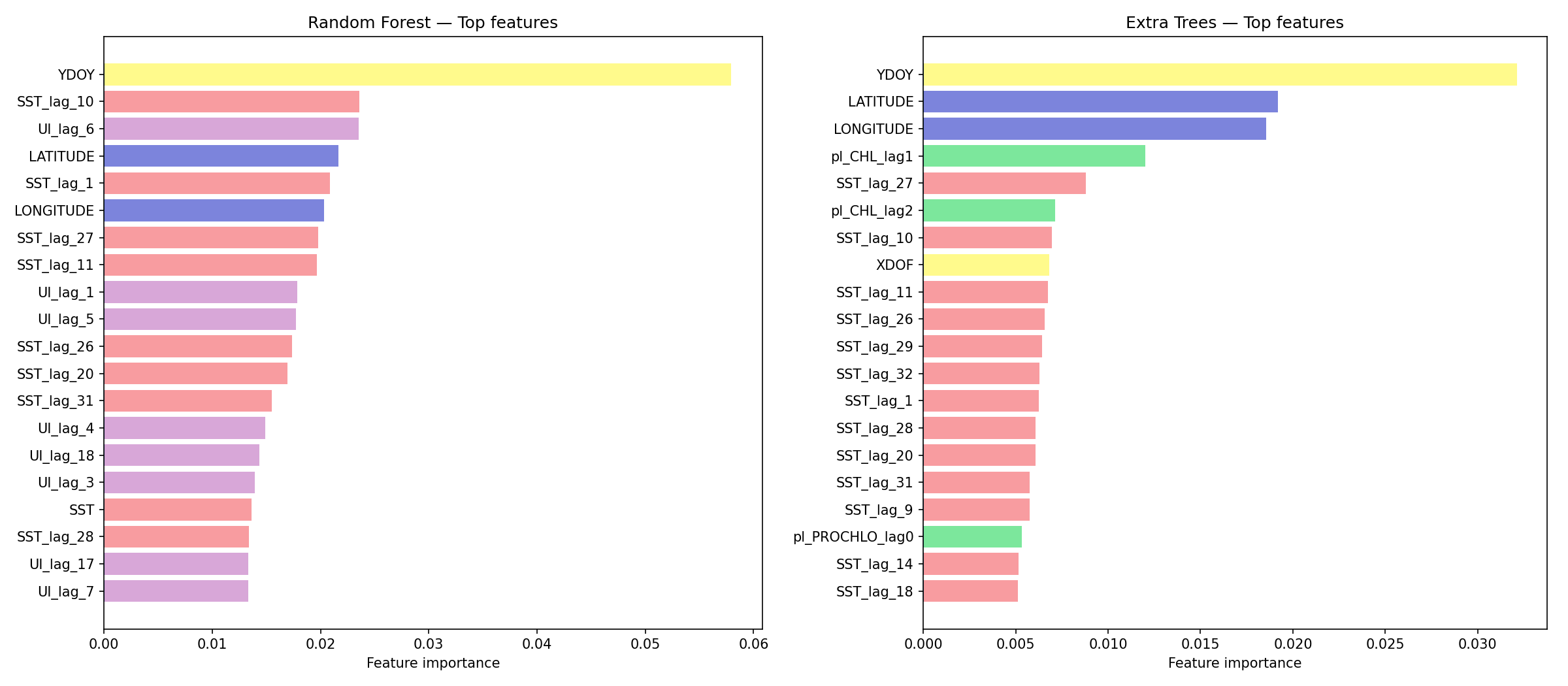}
\caption{Top features for the model using Environmental-only predictors (Random Forest) and Environmental+Biological predictors (Extra Trees). Yellow bars denote temporal indicators (XDOY, YDOY), blue bars denote spatial coordinates (Latitude, Longitude), red bars denote same-day and lagged SST, purple bars denote same-day and lagged UI, and green bars denote biological variables (chlorophyll and plankton functional types).}
\label{fig:importance}
\end{figure}

\begin{figure}[H]
\centering
\includegraphics[width=0.85\textwidth]{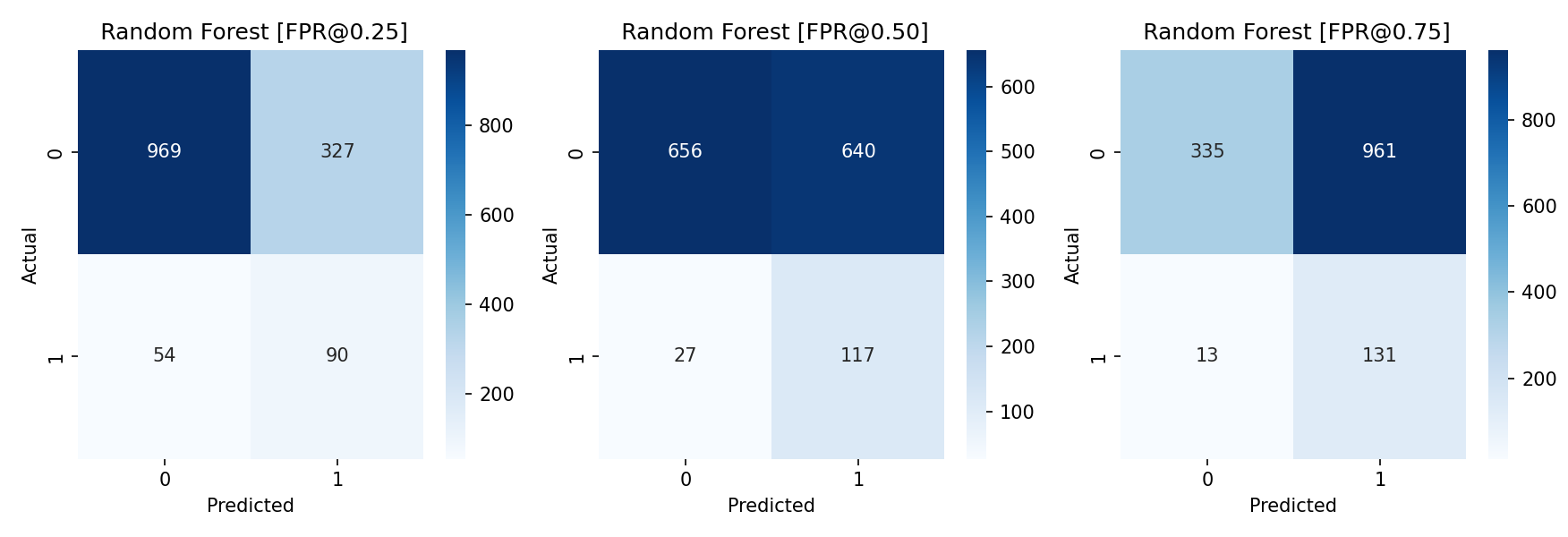}
\caption{Confusion matrices for the Random Forest model (Environmental-only) evaluated at three fixed false-positive rates (0.25, 0.50, 0.75).}
\label{fig:rf_confusion}
\end{figure}

\begin{figure}[H]
\centering
\includegraphics[width=0.85\textwidth]{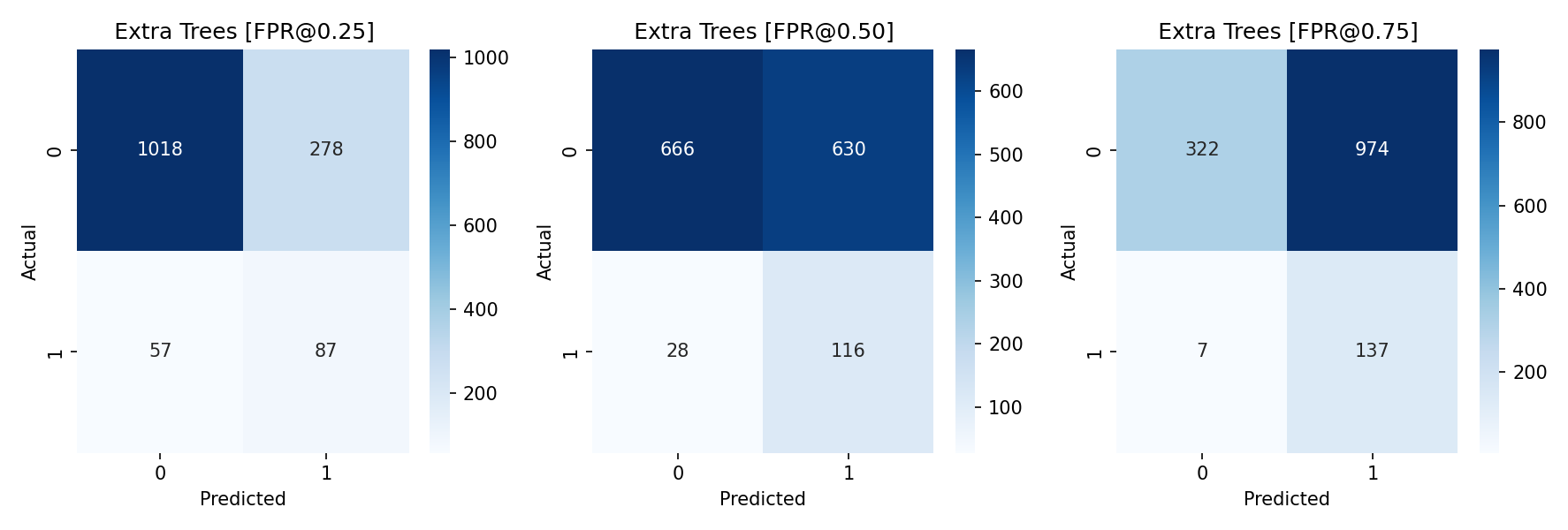}
\caption{Confusion matrices for the Extra Trees model (Environmental+Biological) evaluated at three fixed false-positive rates (0.25, 0.50, 0.75).}
\label{fig:et_confusion}
\end{figure}

\begin{figure}[H]
\centering
\begin{subfigure}[t]{0.45\textwidth}
  \centering
  \includegraphics[width=\textwidth]{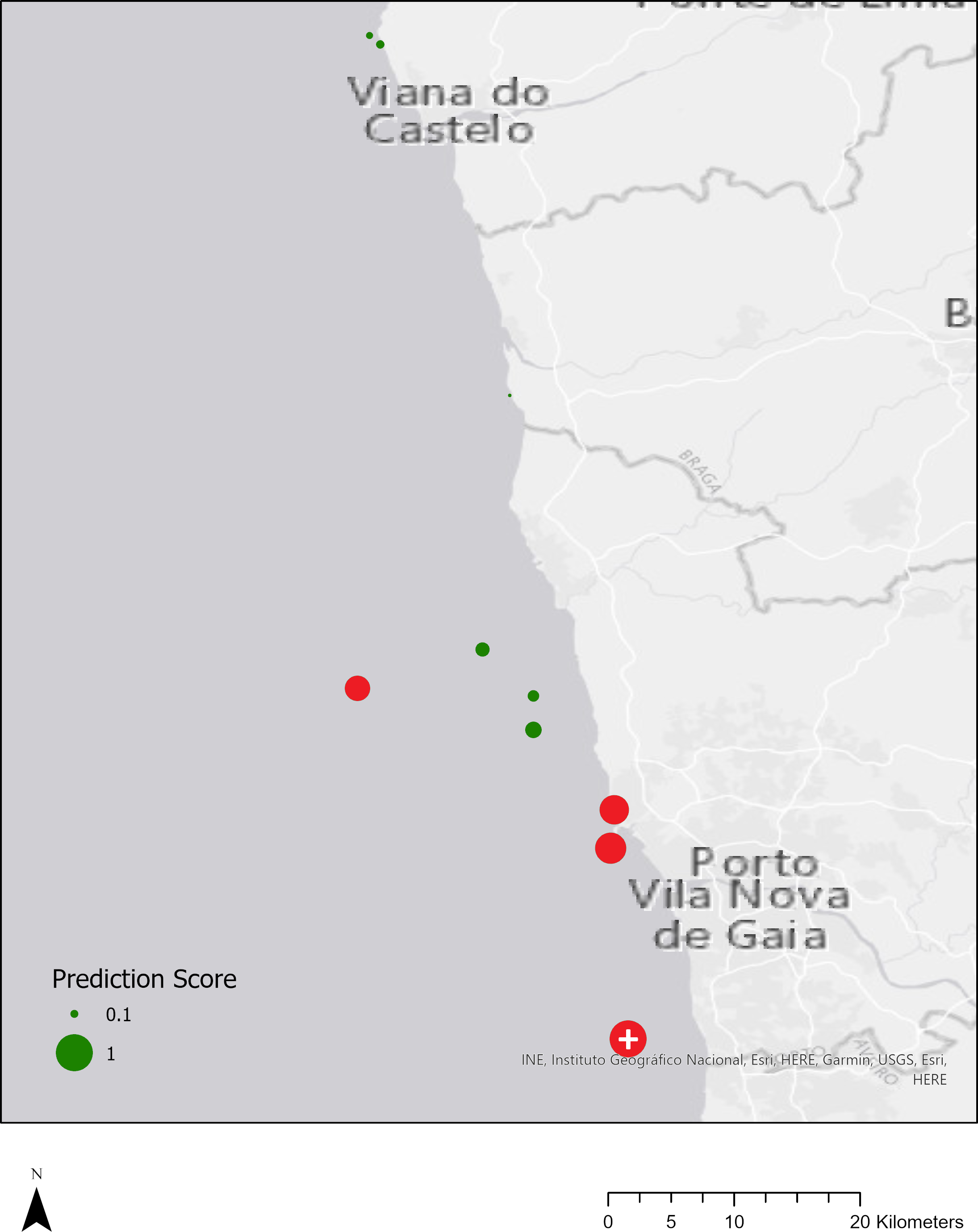}
  \caption{15/09/2022 (event day)}
  \label{fig:hab_spatial_1509}
\end{subfigure}
\hfill
\begin{subfigure}[t]{0.45\textwidth}
  \centering
  \includegraphics[width=\textwidth]{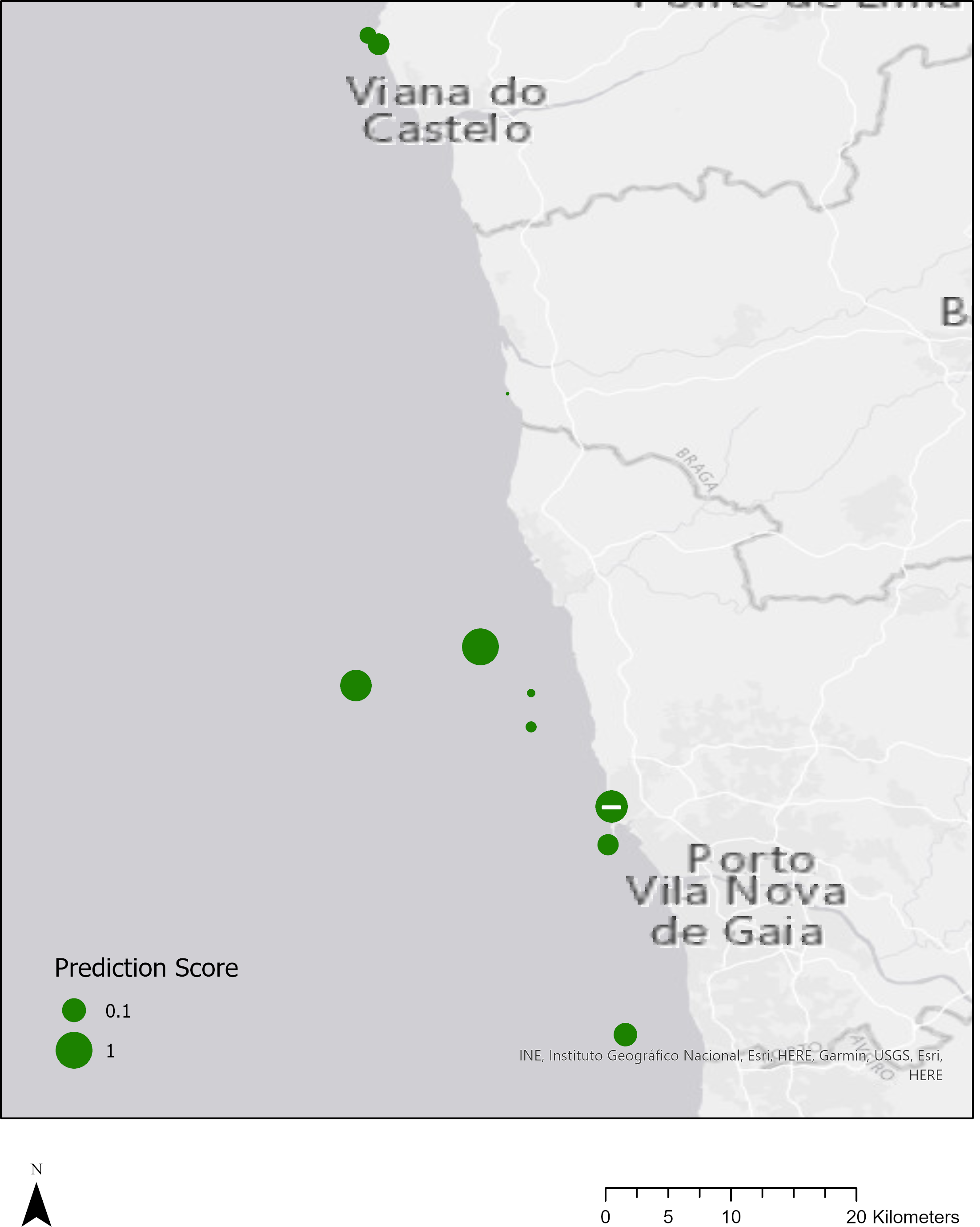}
  \caption{25/09/2022 (no-event day)}
  \label{fig:hab_spatial_2509}
\end{subfigure}
\caption{Spatial illustration of HAB predictions and observations along the Portuguese coast on two contrasting days in September 2022. White symbols at sampled locations denote \textit{in-situ} observations (``$-$'' no HAB; ``$+$'' HAB). Circle size represents the Extra Trees prediction score (larger circles indicate higher predicted risk); colour denotes the predicted class (green: no HAB; red: HAB).}
\label{fig:hab_spatial_panels}
\end{figure}

Beyond overall discrimination metrics, we further examined the internal behaviour and operational characteristics of the best-performing HAB models by analysing feature-importance rankings, decision-threshold confusion matrices, and spatial prediction patterns, as described in Section~\ref{subsec:model_training}. These analyses provide complementary insight into which predictors drive model decisions, how predictive skill translates into false-positive and false-negative trade-offs, and how model outputs can support spatially adaptive monitoring strategies.
\paragraph{Feature importance rankings.}
Figure~\ref{fig:importance} shows the top-ranked predictors for the Random Forest model trained with Environmental-only features. Seasonal indicators (XDOY, YDOY), spatial coordinates, and lagged environmental variables (SST and UI) dominate the importance spectrum, highlighting the role of seasonality, location, and recent physical forcing in HAB discrimination. When biological variables are added, the Extra Trees model exhibits a distinct but related importance structure (Figure~\ref{fig:importance}). While seasonal and spatial predictors remain prominent, lagged Chl-\textit{a} and selected plankton functional types appear among the most influential features, alongside multi-week SST lags. This shift reflects the contribution of biological context in refining HAB predictions beyond purely physical drivers.
\paragraph{Decision thresholds and confusion-matrix analysis.}
To evaluate model behaviour under explicit operational alert constraints, we analysed confusion matrices at three fixed false-positive rates (FPR = 0.25, 0.50, 0.75; Figures~\ref{fig:rf_confusion} and~\ref{fig:et_confusion}). This approach allows direct quantification of the trade-off between missed HAB events and false alerts under increasingly permissive warning strategies, without optimising thresholds post hoc.
For the Random Forest model (Environmental-only), increasing the FPR from 0.25 to 0.50 increased HAB detection from 62.5\% (90 of 144 events) to 81.3\% (117 of 144 events), while reducing the miss rate from 37.5\% to 18.8\%. At the highest alert level (FPR = 0.75), the model captured 91.0\% of HAB events, missing only 13 events, albeit at the cost of a substantially increased false-alert burden. Precision declined monotonically with increasing FPR, reflecting the strong class imbalance inherent to HAB occurrence and the expected trade-off between sensitivity and alert specificity. The Extra Trees model (Environmental+Biological) exhibited similar behaviour at low and moderate false-positive rates but showed a clear advantage under high-sensitivity operating conditions. At FPR = 0.75, the model detected 95.1\% of HAB events (137 of 144), missing only seven events, compared to 13 missed events for the Random Forest. At moderate alert levels (FPR = 0.50), both models achieved comparable recall (approximately 80\%), indicating that the inclusion of biological predictors primarily improves performance in regimes prioritising maximal event capture rather than conservative alerting.
Across both models, these results demonstrate that a substantial fraction of HAB events can be identified under realistic false-positive constraints, with detection rates exceeding 80\% at moderate alert levels and approaching complete capture under permissive thresholds. Rather than defining a single optimal operating point, the confusion-matrix analysis clarifies how model outputs can be aligned with different monitoring objectives, ranging from conservative screening to high-sensitivity early-warning applications.
\paragraph{Spatial prediction patterns.}
Figure~\ref{fig:hab_spatial_panels} illustrates spatial HAB predictions along the Portuguese coast for two representative sampling days in September 2022, a highly sampled month. The two maps are separated by 10 days (15/09/2022 and 25/09/2022) and provide contrasting examples of event versus no-event conditions. In both panels, model output is shown as spatially distributed predictions from the Extra Trees classifier: colour encodes the predicted class (red: HAB; green: no HAB), and proportional symbol size encodes the predicted probability score (larger circles indicate stronger model confidence). White centre symbols at sampled locations denote the \textit{in-situ} classification (``$+$'' HAB; ``$-$'' no HAB), while unsampled locations have no centre marker. Beyond visual validation at sampled locations, these probability maps provide an operational mechanism to prioritise adaptive sampling. Under predicted HAB conditions (red circles), the highest-priority follow-up sampling locations are those with the largest red circles (highest predicted risk), followed by progressively smaller red circles. Conversely, when the model predicts broadly no HAB (only green circles), priority can shift toward uncertainty checking by sampling the smallest green circles first (lowest-confidence ``no-event'' predictions), then progressively larger green circles. In the event-day example (15/09/2022; Figure~\ref{fig:hab_spatial_panels}a), the next sampling priority is indicated in the cluster immediately north of the sampled location, where the model assigns elevated predicted risk at an unsampled site. In the no-event example (25/09/2022; Figure~\ref{fig:hab_spatial_panels}b), the analogous ``next check'' priority occurs again in the adjacent northern cluster, where the smallest green symbols identify the lowest-confidence no-event predictions.

\section{Discussion}
\subsection{Predictability of \textit{Pseudo-nitzschia} HABs and ecological drivers}
This study demonstrates that the occurrence of PN harmful algal blooms along the Portuguese Atlantic coast can be predicted with moderate but robust skill (ROC--AUC $\approx$ 0.70--0.77) using only satellite-derived environmental and biological variables under strict spatio-temporal validation. Given the strong physical heterogeneity of the Iberian upwelling system and the exclusion of \textit{in situ} biological predictors, this level of discrimination indicates that remotely sensed proxies capture a meaningful fraction of the environmental window conducive to bloom development.
Ensemble tree models consistently outperformed linear and kernel-based classifiers, highlighting the importance of non-linear interactions among seasonality, location, physical forcing, and biological state. Similar performance ranges have been reported in other regional HAB forecasting studies that focus on bloom presence. For example, \textcite{lane2009logistic} showed that environmental variables could successfully discriminate PN bloom conditions in Monterey Bay, while \textcite{Harris2021FL} and \textcite{Pan2021YellowSea} reported ROC--AUC values of approximately 0.70--0.74 for Random Forest and gradient-boosted models predicting HABs in coastal Florida and the Yellow Sea, respectively. The present results therefore align with, and modestly extend, the performance conditions reported for comparable coastal systems. The inclusion of biological predictors (Chl-\textit{a} and PFTs) yielded a consistent improvement in HAB discrimination for ensemble models, with Extra Trees achieving the highest overall performance (ROC--AUC = $0.77 \pm 0.06$). This result supports the view that biological indicators refine bloom likelihood estimates by constraining predictions to periods of enhanced phytoplankton biomass and favourable ecosystem structure, rather than by directly identifying \textit{Pseudo-nitzschia} itself. Such behaviour is consistent with previous work showing that Chl-\textit{a} anomalies and PFT-derived indicators act as early signals of bloom-favourable regimes \parencite{Zhang2018PRD, Wilson2020NWAtlantic}.
Feature-importance analyses provide further ecological insight into the drivers of model predictions. Seasonal indicators, particularly the cosine of day-of-year, were the most influential predictors across all models, reaffirming the dominant phenological control on PN blooms in the Iberian upwelling system. Spatial coordinates also ranked highly, reflecting persistent alongshore gradients in hydrography, riverine influence, and anthropogenic pressure that are not fully resolved by the predictor set but nevertheless shape bloom susceptibility. Lagged physical variables, especially sea surface temperature at time scales of one to four weeks, were consistently among the most important environmental predictors. The prominence of multi-week SST lags suggests that bloom development reflects cumulative thermal conditions and growth history rather than instantaneous forcing, consistent with known diatom growth and accumulation timescales. The importance of lagged upwelling indices further supports the role of pulsed nutrient enrichment followed by relaxation and retention phases in initiating bloom conditions.
When biological predictors were included, short-lag Chl-\textit{a} emerged as a key feature, indicating that recent increases in bulk phytoplankton biomass act as strong precursors to HAB occurrence. Among plankton functional types, prochlorophytes appeared more frequently than diatom-specific PFTs in the top-ranked features, a pattern that likely reflects the optical environment of the nearshore sampling stations rather than a direct ecological association with \textit{Pseudo-nitzschia}. In optically complex coastal waters subject to strong riverine influence, satellite retrievals of small-cell phytoplankton groups tend to be more stable than diatom-specific signals, and prochlorophyte abundance may act as a proxy for the broader productivity regime rather than community composition \textit{per se} \parencite{brewin2017PFTReview}.
\subsection{Model performance, validation strategy, and operational implications}
A key strength of this study is the leakage-resistant spatio-temporal validation framework. As described in Section~\ref{subsec:spatio_temporalCV}, the evaluation explicitly removes short-term temporal persistence and spatial autocorrelation as sources of predictive skill. This ensures that model performance reflects genuine generalisation to unseen environmental conditions and coastal regimes. Under this conservative protocol, the achieved performance (ROC--AUC up to $0.77 \pm 0.06$) should be interpreted as a realistic estimate of forecasting capability rather than an upper-bound accuracy. In contrast, higher scores frequently reported in the literature are typically obtained using random or weakly structured validation schemes, which allow information leakage across nearby observations and can lead to over-optimistic assessments of model skill. Comparisons with recent work highlight the importance of validation strategy when interpreting reported model performance. Studies reporting higher accuracies or AUC values often rely on dense \textit{in situ} sensor networks or relaxed validation schemes \parencite{Bertani2021RFvsSVM, Chen2022LakeErie}. From an operational perspective, a model that achieves moderate discrimination under strict out-of-distribution validation is more valuable than one achieving near-perfect accuracy under leakage-prone conditions. The former reflects transferable predictive structure, while the latter often captures site-specific or temporally persistent patterns that do not generalise to new events or locations.
Moreover, the translation of discrimination skill into operational decisions requires explicit consideration of false-positive and false-negative trade-offs. The confusion-matrix analyses at fixed false-positive rates demonstrate that the best-performing models can capture the majority of bloom events while maintaining controllable false-alert levels. This flexibility allows decision thresholds to be tuned according to monitoring capacity, economic cost, and risk tolerance. The spatial prediction maps further illustrate the potential of probabilistic model outputs to support adaptive monitoring. By highlighting unsampled locations with elevated predicted risk or high classification uncertainty, the models provide a mechanism for prioritising follow-up sampling in space and time. Rather than replacing existing \textit{in situ} programmes, such tools can enhance their efficiency by directing effort toward locations where additional information is most valuable. Overall, these results indicate that satellite-driven, machine-learning-based HAB forecasting is feasible at regional scales when combined with biologically informed predictors and rigorous validation. While important limitations remain, particularly for toxin prediction, the demonstrated skill for HAB occurrence supports the integration of such models into adaptive, risk-informed monitoring frameworks along the Portuguese coast.
\subsection{Practical implementation in monitoring laboratories}
A recurrent challenge in HAB forecasting research is the gap between demonstrating predictive skill and delivering a tool that a monitoring laboratory can operate routinely. The personnel responsible for shellfish safety assessments are predominantly biologists and ecotoxicologists, not data scientists, and the question of where to begin with a satellite-driven model is a genuine barrier even when the underlying results are strong. The observational inputs required are all freely accessible: sea surface temperature from the CMEMS SST product, the upwelling index derived from the same wind-stress fields used in this study, and the biological predictors (chlorophyll-\textit{a} and plankton functional types) from the CMEMS North Atlantic Ocean Colour BGC product. Both pipelines are implemented in the open-source \texttt{CMEMS\_Data\_Analysis} repository, which automates download, spatial collocation with fixed monitoring station coordinates, and temporal lag construction. In practice, a laboratory technician would run a single script each morning to update the predictor table for all active sampling locations with the previous day's satellite data. The model itself, once trained and exported, requires no specialist infrastructure to apply: a trained Extra Trees classifier can be serialised and loaded in a standard Python environment with \texttt{scikit-learn}, and predicted probabilities for each monitoring location can be written to a CSV file readable in any spreadsheet application. A simple dashboard, even a colour-coded spreadsheet updated daily, could display the predicted bloom risk score for each station alongside the alert threshold selected by the monitoring agency. What remains an unsolved problem, and one that the present results do not resolve, is the institutional pathway from a validated research model to a sanctioned operational tool. The framework developed here, including the clustering algorithm, the cross-validation design, and the trained classifiers, is intended as a documented and reproducible starting point for that process rather than a finished product.
\section{Conclusion}
Predicting when and where \textit{Pseudo-nitzschia} harmful algal blooms will occur is an inherently difficult problem. Bloom development emerges from the interaction of seasonal forcing, local physical dynamics and community-level biological state, and the conditions that ultimately matter for public health remain only partially visible in satellite observations. The central argument of this study is that satellite-driven machine learning can already deliver operationally useful predictive skill for HAB occurrence when two conditions are met: the evaluation protocol is physically honest, and the predictor set is biologically informed.
The primary contribution of this work is not the performance figures themselves but the framework that produces them. By combining a river-aware spatial clustering strategy with a cross-validation scheme that withholds entire calendar years and spatial clusters simultaneously, we constructed an evaluation protocol that prevents both temporal and spatial leakage and closely reflects the extrapolative challenge of real-world forecasting. Under these constraints, the Extra Trees classifier incorporating satellite-derived biological predictors alongside environmental forcing achieved a ROC--AUC of $0.77 \pm 0.06$, a result that should be read as a conservative and operationally realistic baseline rather than an optimistic accuracy figure. The ecological structure underlying these results is coherent with what is known about \textit{Pseudo-nitzschia} dynamics in eastern boundary upwelling systems: seasonal timing, spatial position and multi-week histories of sea surface temperature and upwelling together account for the dominant share of discriminative skill, while short-lag chlorophyll-\textit{a} provides a meaningful biological precursor signal.
From an operational perspective, the probabilistic spatial outputs generated by the Extra Trees classifier provide a concrete mechanism for adaptive monitoring. Adjusting the alert threshold from FPR 0.25 to FPR 0.50 raises HAB detection from 62.5\% to 81.3\% for the environmental model, while the biologically enriched Extra Trees model reaches 95.1\% event capture at FPR 0.75. This trade-off surface provides a quantitative basis for threshold selection that can be tailored to the monitoring capacity and risk tolerance of national agencies such as IPMA. Rather than replacing existing \textit{in situ} programmes, the framework is designed to complement them: satellite-driven model outputs identify when and where conditions are conducive to \textit{Pseudo-nitzschia} proliferation, while \textit{in situ} measurements provide the toxin confirmation required for management decisions.
\paragraph{Limitations and future work.}
Three limitations bound the present results and define the natural next steps. (i) The supervised target is HAB \emph{occurrence}; advancing to \emph{toxicity} prediction will require a differently structured dataset in which shellfish toxin measurements are matched to individual sampling events rather than to ban intervals. (ii) The fixed 0--45 day tabular lag representation encodes temporal structure only implicitly; sequence-aware architectures such as temporal fusion transformers, FT-Transformer or TabPFN \parencite{Lim2021TFT, Gorishniy2021FTTransformer, Hollmann2022TabPFN, Wu2021Autoformer} could capture longer-range dependencies and regime shifts, provided that the physically informed cross-validation design introduced here is preserved \parencite{jolt2024, timeseriesgym2025}. (iii) The framework is calibrated on the L1--L2 hotspot; extension to the central and southern production zones will require re-deriving the river-aware clustering and re-evaluating dominant lag structures under different upwelling regimes \parencite{cruz2021review, yu2021predicting}. Despite these limitations, the present study demonstrates that satellite-only \textit{Pseudo-nitzschia} HAB forecasting is feasible at operationally useful skill levels along the Portuguese Atlantic coast, and that evaluation integrity (not model complexity) is what makes such estimates reliable. By making the spatio-temporal validation framework and the river-aware clustering algorithm openly available, this work aims to contribute to the broader goal of satellite-supported HAB early-warning systems in eastern boundary upwelling environments.
\section*{Acknowledgements}
The authors gratefully acknowledge the Instituto Português do Mar e da Atmosfera (IPMA) for making available online the long-term phytoplankton monitoring records and shellfish harvesting-ban data used in this study, and the Copernicus Marine Environment Monitoring Service (CMEMS) and the European Marine Observation and Data Network (EMODnet) for the satellite and coastal contaminant datasets used as predictor variables. We thank colleagues at ISR -- Institute for Systems and Robotics, at MARETEC -- Marine, Environment and Technology Centre (Instituto Superior Técnico, Universidade de Lisboa), and from IPMA--CIIMAR -- Interdisciplinary Centre of Marine and Environmental Research for fruitful discussions during the development of the predictive framework. This work was supported by Fundação para a Ciência e a Tecnologia (FCT), Portugal,
through LARSyS (DOI:~10.54499/LA/P/0083/2020, 10.54499/UIDP/50009/2020,
and 10.54499/UIDB/50009/2020) and through the PhD scholarship of A.~Bnoussaad
(UI/BD/155012/2023). L.~Pinto was supported by national funds through FCT --
Fundação para a Ciência e a Tecnologia, I.P., under the FCT-Tenure programme
(Reference 2023.15700.TENURE.008).
\section*{CRediT author contribution statement}
\textbf{Ayman Bnoussaad:} Conceptualisation, Methodology, Software, Formal analysis, Investigation, Data curation, Writing -- original draft, Visualisation.
\textbf{El Khalil Cherif:} Conceptualisation, Writing -- review \& editing, Co-Supervision.
\textbf{Ligia Pinto:} Writing -- review \& editing.
\textbf{Ramiro Neves:} Writing -- review \& editing.
\textbf{Alexandra D. Silva:} Data curation, Resources, Writing -- review \& editing.
\textbf{Alexandre Bernardino:} Conceptualisation, Methodology, Writing -- review \& editing, Supervision.
\section*{Declaration of competing interest}
The authors declare that they have no known competing financial interests or personal relationships that could have appeared to influence the work reported in this paper.
\section*{Declaration of generative AI and AI-assisted technologies in the writing process}
During the preparation of this work the authors used Claude (Anthropic) in
order to improve readability and language of selected passages. After using
this tool, the authors reviewed and edited the content as needed and take
full responsibility for the content of the publication.
\section*{Data and code availability}
The IPMA monitoring records used in this study are available online from the Instituto Português do Mar e da Atmosfera\footnote{IPMA shellfish bivalve monitoring portal, \href{https://www.ipma.pt/pt/bivalves/index.jsp}{https://www.ipma.pt/pt/bivalves/index.jsp}.}. Satellite predictors were obtained from the Copernicus Marine Service\footnote{Copernicus Marine Service data portal, \href{https://data.marine.copernicus.eu}{https://data.marine.copernicus.eu}.} and processed using the open-source \texttt{CMEMS\_Data\_Analysis} repository\footref{fn:cmems}. The river-aware clustering algorithm is provided in the \texttt{ROFI\_Clustering} repository\footnote{\label{fn:rofi}Bnoussaad, A. (2026), ``ROFI\_Clustering,'' \href{https://github.com/aymansvvd/ROFI_Clustering}{GitHub repository}.}. Trace-metal data are from EMODnet Chemistry (v2024)\footref{fn:emodnet}.
\appendix
\section{River-aware spatial clustering methodology}
\label{app:rofi_clustering}
This appendix provides the full methodological description of the river-aware spatial clustering approach used to define coastal segmentation in the main analysis. For the segmentation shown in Figure~\ref{fig:river_clusters}, we employed the density-based spatial clustering algorithm DBSCAN with an augmented distance metric incorporating freshwater influence. Geographic distances between sampling locations were computed using the haversine formula, yielding great-circle distances in kilometres. DBSCAN identifies spatial clusters as regions of high point density separated by areas of lower density, without requiring the number of clusters to be specified a priori. To incorporate freshwater influence, the pairwise distance metric was augmented to include similarity in river exposure. For each sampling location $i$, the haversine distance $d_i$ to its nearest river mouth was computed. River mouths were assigned weights proportional to stream order, which serves as a proxy for relative discharge magnitude. The distance between two sampling locations $i$ and $j$ was then defined as
\[
D_{ij} = \sqrt{D_{ij}^{\text{geo}\,2} +
\left( \alpha \, |d_i - d_j| \right)^2 },
\]
where $D_{ij}^{\text{geo}}$ is the haversine geographic distance (km) between the two locations, $d_i$ and $d_j$ are their respective distances to the nearest river mouth, and $\alpha$ is a weighting parameter controlling the contribution of freshwater-exposure similarity to the clustering metric. This formulation ensures that locations geographically close but influenced by different river systems are less likely to be assigned to the same spatial cluster. DBSCAN was parameterised using a neighbourhood radius $\varepsilon$ (km) and a minimum number of samples per cluster. Noise points were subsequently reassigned to their nearest non-noise cluster to ensure spatial completeness of the coastal segmentation. The resulting solution yielded six coherent coastal clusters, representing distinct river-influenced spatial regimes along the northern Portuguese coast. This unsupervised algorithm is particularly appropriate in this context because (i) coastal sampling density is spatially irregular, (ii) cluster shapes may follow coastline geometry rather than forming convex partitions, and (iii) the true number of spatial regimes is not known a priori.
The \texttt{ROFI\_Clustering} framework also provides a $k$-means implementation for applications where a fixed number of clusters is desired. In that case, latitude, longitude, and a river-influence variable defined as
\[
I_i = \frac{w_{r^*}}{1 + d_i / L},
\]
with $L=30$~km, a decay scale consistent with reported ROFI extents on eastern boundary and river-dominated continental shelves \parencite{kristiansen1997phytoplankton,simionato2006rioplata,dodrill2022columbia}, are standardised and combined into an augmented feature vector prior to clustering. The number of clusters $K$ may either be prescribed or selected automatically by maximising the silhouette score. Compared to DBSCAN, $k$-means is preferable when: (i) a predetermined number of spatial blocks is required (e.g., for cross-validation design), (ii) cluster compactness and global partitioning are desired, or (iii) sampling density is approximately uniform. The river-aware clusters derived here provide a physically grounded coastal segmentation that captures both spatial proximity and freshwater forcing structure. These clusters are subsequently used to define spatial blocking units in the spatio-temporal cross-validation framework, ensuring that predictive model evaluation reflects realistic extrapolation across distinct river-influenced coastal regimes rather than random spatial mixing.
The clustering workflow was implemented in a reproducible Python-based framework (\texttt{ROFI\_Clustering})\footref{fn:rofi}, designed to integrate geographic sampling data with river network information to approximate Regions of Freshwater Influence (ROFIs). The workflow takes as input (i) sampling site coordinates and (ii) river mouth locations, from which nearest-river distances and associated exposure metrics are computed for each site. In addition to cluster assignments, the framework produces summary statistics of cluster composition, site-to-river distance metrics, and geospatial visualisations to support interpretation of the resulting coastal segmentation. Both DBSCAN and $k$-means clustering algorithms are supported within the same framework, allowing comparison between density-based and partition-based approaches under consistent feature representations.
\section{CMEMS data retrieval and preprocessing}
\label{app:cmems_pipeline}
All environmental and biological predictors used in this study were derived from Copernicus Marine Environment Monitoring Service (CMEMS) products using a unified and reproducible data-processing pipeline (\texttt{CMEMS\_Data\_Analysis}). The pipeline was designed to automate data access, ensure consistent spatio-temporal alignment with \textit{in situ} observations, and generate analysis-ready predictors for machine-learning applications.
\paragraph{Upwelling-index derivation.}
The upwelling index was derived from surface wind stress following classical Ekman theory \parencite{ekman1905motion}. Let $u$ and $v$ denote the zonal and meridional wind components, and let $|\mathbf{U}|=\sqrt{u^{2}+v^{2}}$ be the wind speed. Wind stress was computed using a bulk aerodynamic formulation,
\begin{equation}
\tau_{x} = \rho_{\text{air}}\,C_{D}\,|\mathbf{U}|\,u, \qquad
\tau_{y} = \rho_{\text{air}}\,C_{D}\,|\mathbf{U}|\,v,
\end{equation}
where $\rho_{\text{air}}$ is air density and $C_{D}$ is the drag coefficient. The Coriolis parameter is $f = 2\,\Omega\,\sin\varphi$, with $\Omega$ the Earth's rotation rate and $\varphi$ latitude. Following the coastal upwelling formulation of \textcite{bakun1973coastal}, the UI was computed as the cross-shore Ekman transport projected onto the local coastline orientation,
\begin{equation}
\mathrm{UI}
= \frac{\tau_{x}\,\sin\theta - \tau_{y}\,\cos\theta}
       {\rho_{\text{water}}\,f},
\end{equation}
where $\theta$ is the coastline orientation (degrees from east) and $\rho_{\text{water}}$ is seawater density.
\paragraph{Data pipeline.}
CMEMS data were accessed programmatically through a Python-based workflow interfacing with the Copernicus Marine data infrastructure. For each environmental and biological variable, daily Level-3 NetCDF files were retrieved over the spatial domain encompassing the Portuguese coast, using a modular design that facilitates reproducibility and extension to additional datasets. All data were handled in NetCDF format using \texttt{xarray}, \texttt{netCDF4}, and \texttt{numpy}. Preprocessing steps included extraction of relevant variables, harmonisation of coordinate conventions (e.g., longitude), conversion of time variables to absolute timestamps, and unit standardisation (e.g., conversion of SST from Kelvin to degrees Celsius). Environmental and biological variables were matched with \textit{in situ} sampling observations using a nearest-neighbour matching strategy in both space and time. For each sampling event, the corresponding daily CMEMS file was identified and the closest valid grid cell selected within a spatial tolerance of $0.05^{\circ}$, with temporal alignment enforced at the daily scale. Observations without a valid satellite match within this tolerance were assigned missing values (\texttt{NaN}) to avoid introducing artificial interpolation. Quality control procedures included validation of spatial matches relative to coastline boundaries, detection of missing or corrupted files, and propagation of missing values where observations were unavailable. No spatial or temporal interpolation was applied at this stage, preserving the integrity of the original satellite products. Lagged versions of the main predictors were constructed by extracting historical values up to 45 days prior to each sampling date; lag construction strictly respects temporal causality, ensuring that only information available at or before the sampling time is included.
All processing steps were implemented within the open-source Python package \texttt{CMEMS\_Data\_Analysis}\footref{fn:cmems}, which provides modular components for data retrieval, NetCDF processing, spatio-temporal collocation, and lagged feature construction, and is fully reproducible and extensible to additional variables, regions, and temporal resolutions.
\vspace{6pt}

\printbibliography
\end{document}